\documentclass[11pt]{article}

\usepackage[final]{acl}

\usepackage{times}
\usepackage{latexsym}

\usepackage[T1]{fontenc}

\usepackage[utf8]{inputenc}
 
\usepackage{microtype}

\usepackage{inconsolata}

\usepackage{graphicx}
\usepackage{pifont}
\usepackage{makecell}
\usepackage{booktabs}
\usepackage{multirow}

\usepackage{enumitem}
\usepackage{url}
\usepackage{dashrule}
\usepackage[skins]{tcolorbox}

\usepackage{colortbl}
\usepackage{xcolor}
\PassOptionsToPackage{table, xcdraw}{xcolor} 

\usepackage{afterpage}
\usepackage{threeparttable}

\usepackage{cuted}

\usepackage{amsmath}
\usepackage{caption}
\usepackage{array}
\usepackage{wrapfig}
\usepackage{adjustbox}
\usepackage{amsfonts}       
\usepackage{nicefrac}       
\usepackage{hyperref}
\usepackage{url}
\newcommand{\up}[1]{\textcolor{red}{\scriptsize{+#1}}}
\newcommand{\down}[1]{\textcolor{table-green}{\scriptsize{#1}}}
\definecolor{spa-blue}{RGB}{54,130,190}
\definecolor{spa-green}{RGB}{69, 167, 118}
\definecolor{spa-yellow}{RGB}{238, 195, 84}
\definecolor{spa-orange}{RGB}{240, 83, 38}
\definecolor{spa-pink}{RGB}{223, 56, 129}
\definecolor{spa-purple}{RGB}{132, 75, 179}
\definecolor{table-green}{RGB}{29, 150, 28}
\definecolor{table-red}{RGB}{221, 68, 71}
\definecolor{table-purple}{RGB}{226, 226, 250}
\definecolor{text-green}{RGB}{84, 130, 53}
\definecolor{text-red}{RGB}{192, 0, 0}
\definecolor{table-yellow}{RGB}{255, 255, 204}
\definecolor{category-spatial}{RGB}{239, 176, 167}
\definecolor{category-science}{RGB}{155, 201, 251}
\definecolor{category-location}{RGB}{251, 237, 197}
\definecolor{category-daily}{RGB}{162, 162, 240}
\definecolor{background-blue}{RGB}{221, 231, 250}
\definecolor{background-grey}{HTML}{EAEAEA}
\definecolor{nvidiagreen}{RGB}{119,185,0}

\title{
        \raisebox{-0.3em}{\includegraphics[height=1.6em]{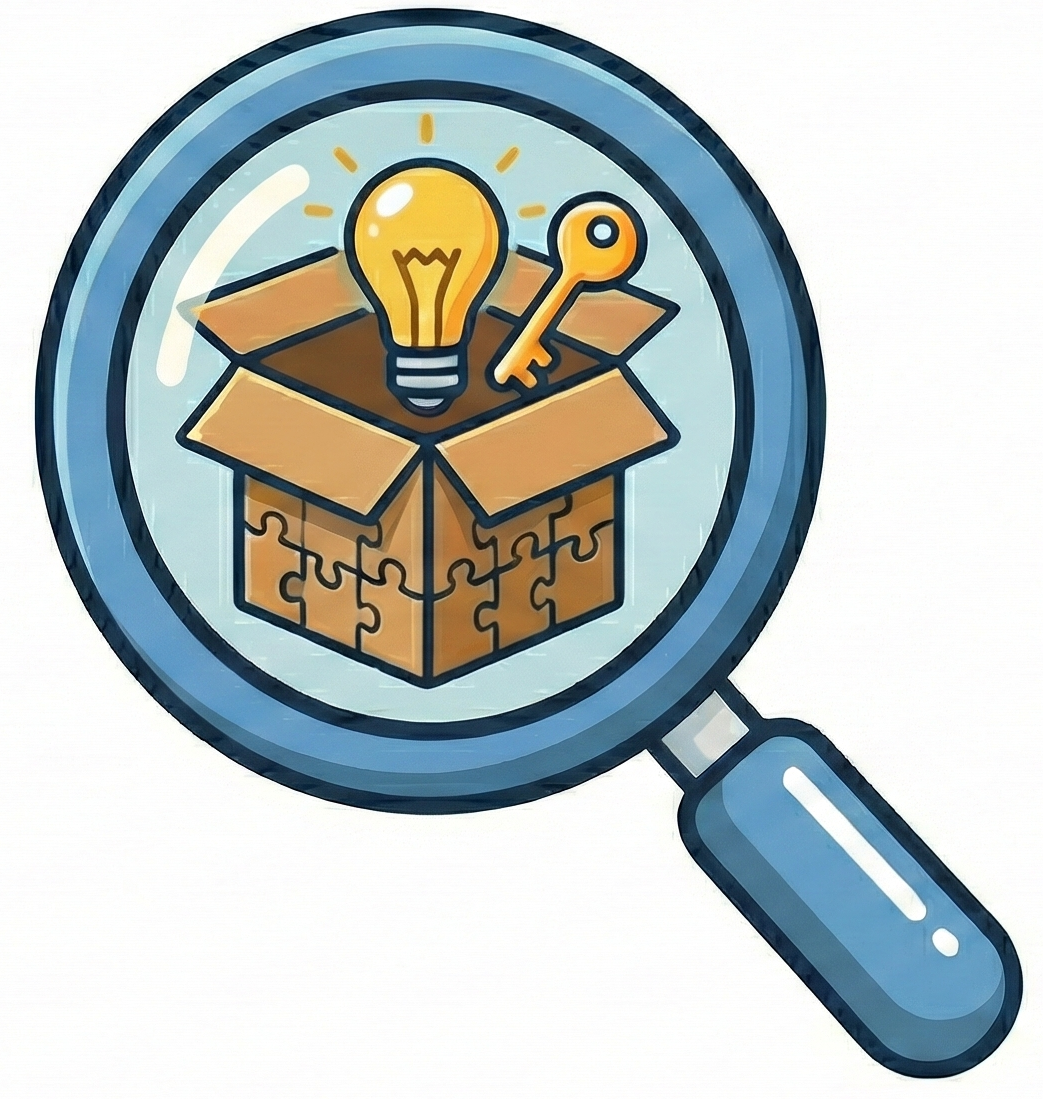}} Seek-and-Solve: Benchmarking MLLMs \\ for Visual Clue-Driven Reasoning in Daily Scenarios}

\author{
 \textbf{Xiaomin Li~\textsuperscript{1~$\star$}},
 \textbf{Tala Wang~\textsuperscript{1~$\star$}},
 \textbf{Zichen Zhong~\textsuperscript{1~$\star$}},
 \textbf{Ying Zhang~\textsuperscript{2}},
 \\
 \textbf{Zirui Zheng~\textsuperscript{1}},
 \textbf{Takashi Isobe~\textsuperscript{3}},
 \textbf{Dezhuang Li~\textsuperscript{1}},
 \textbf{Huchuan Lu~\textsuperscript{1}},
 \textbf{You He~\textsuperscript{3}},
 \textbf{Xu Jia~\textsuperscript{1~$\dagger$}}
\\
 \textsuperscript{1}~Dalian University of Technology,
 \textsuperscript{2}~WeChat, Tencent Inc.,
 \textsuperscript{3}~Tsinghua University
\\
xmli22@mail.dlut.edu.cn, xjia@dlut.edu.cn 
\\
\href{https://xiaominli1020.github.io/DailyClue/}{\textcolor{magenta}{https://xiaominli1020.github.io/DailyClue/}}
}

\begin{document}
\maketitle
\let\thefootnote\relax\footnotetext{$^\star$ Equal contribution. This work was done when Xiaomin Li was an intern at WeChat.} 
\let\thefootnote\relax\footnotetext{$^\dagger$ Corresponding author.
}

\begin{abstract}
Daily scenarios are characterized by visual richness, requiring Multimodal Large Language Models~(MLLMs) to filter noise and identify decisive visual clues for accurate reasoning. Yet, current benchmarks predominantly aim at evaluating MLLMs' pre-existing knowledge or perceptual understanding, often neglecting the critical capability of reasoning.
To bridge this gap, we introduce DailyClue, a benchmark designed for visual clue-driven reasoning in daily scenarios. Our construction is guided by two core principles: (1) strict grounding in authentic daily activities, and (2) challenging query design that necessitates more than surface-level perception. Instead of simple recognition, our questions compel MLLMs to actively explore suitable visual clues and leverage them for subsequent reasoning. To this end, we curate a comprehensive dataset spanning four major daily domains and 16 distinct subtasks.
Comprehensive evaluation across MLLMs and agentic models underscores the formidable challenge posed by our benchmark. Our analysis reveals several critical insights, emphasizing that the accurate identification of visual clues is essential for robust reasoning.

\end{abstract}
\section{Introduction}
In recent years, we have witnessed a significant flourish of multimodal large language models (MLLMs)~\cite{gpt5,gemini,qwen3-vl,llava-ov-1.5,internvl3}. These models exhibit strong capabilities in understanding and reasoning, allowing humans to readily employ them as comprehensive encyclopedic sources for knowledge retrieval or versatile creative assistants for content generation. 
The rapid development of MLLMs has consequently stimulated the emergence of diverse benchmarks~\cite{hrbench,ocrbench,mmmu,mmmu-pro,MathVision}, through which the capability limitations of existing models can be systematically evaluated and revealed. 

Existing general visual question answering~(VQA) benchmarks~\cite{mmbench,mme-realworld,realworldqa} primarily focus on simple, explicit factual queries—such as object counting or recognition—that probe only the basic perceptual capabilities~\cite{tir-bench}.
Although the emergence of GPT-o3 has inspired a wave of ``Think-with-Image'' agents~\cite{deepeyes,thyme,pixel_reasoner,treebench,tir-bench,revpt} and datasets~\cite{seekworld,hrbench,v_star,vlm-r3,treebench} capable of utilizing external tools (e.g., zooming, code execution), these agents are typically optimized through reinforcement learning~(RL) to master complex reasoning trajectories~\cite{uni-dpo,reneg}. However, these models still face significant challenges in daily scenarios characterized by visual distractions. In such complex environments, accurate reasoning depends on identifying the correct visual clues amidst the noise. Unfortunately, current benchmarks fail to sufficiently evaluate this critical ability.

\begin{figure*}[!t]
\centering
\includegraphics[width=1\textwidth]{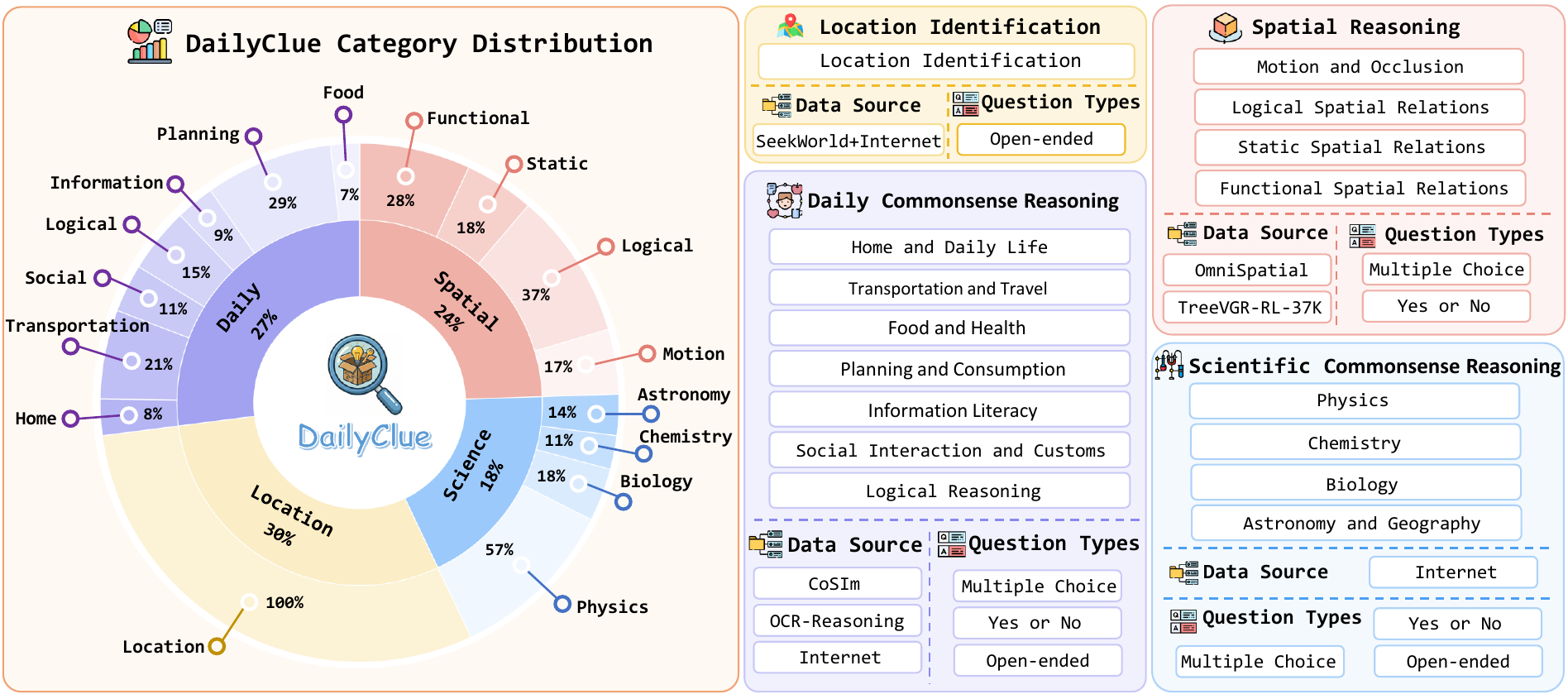}
\caption{Overview of DailyClue. The left panel shows the hierarchical distribution~(labels abbreviated for clarity), while the right panels detail the full taxonomy, data sources, and question types.}
\label{fig:statistics}
\end{figure*}

Moreover, performance on existing benchmarks is reaching a saturation point (exceeding 90\% accuracy)~\cite{deepeyes,v_star,hrbench}, which makes it increasingly difficult to differentiate between top-tier models. This necessitates a more challenging benchmark, specifically tailored to rigorously assess deep reasoning capabilities in daily-life contexts.

In this work, we introduce \textbf{DailyClue}, a comprehensive benchmark designed to evaluate MLLMs' understanding of daily-life scenarios and reasoning grounded in visual clues.
Unlike existing benchmarks that predominantly measure a model's internal world knowledge, DailyClue is strictly \textit{clue-driven} rather than \textit{knowledge-driven}. It isolates a model’s ability to identify and utilize specific visual clue, preventing models from relying on superficial correlations or memorized priors to ``guess'' answers.
DailyClue is guided by two core principles: (1) all scenes are strictly derived from real-world contexts, aligning with the core requirements of embodied AI and multimodal agents to solve multi-step workflows grounded in visual evidence~\cite{agentvista}; (2) questions are non-trivial and cannot be answered through direct visual inspection, instead necessitating deduction from implicit visual clues.
To construct the benchmark, we employ a carefully controlled generation and filtering pipeline. First, top-tier MLLMs generate candidate question–clue–answer triplets. Then, these candidates are verified by multiple peer models, ensuring that only instances remaining consistently challenging are retained. Following rigorous human verification, the final benchmark comprises 666 curated triplets spanning four broad categories and sixteen distinct subtasks, covering multiple-choice, open-ended QA, and binary decision tasks.

We extensively benchmark 25 MLLMs on DailyClue, deriving three critical insights:
\begin{itemize}
    \item DailyClue poses a significant challenge to both humans and top-tier MLLMs, serving as a robust and discriminative testbed for assessing holistic comprehension of daily scenarios.
    \item Model performance is critically bottlenecked by inaccurate visual clue prediction during CoT reasoning; notably, injecting ground-truth clues yields substantial gains.
    \item Explicitly prompting models to actively seek visual clues improves accuracy, yet this benefit exhibits diminishing returns as models' intrinsic reasoning capabilities scale up.
\end{itemize}
\section{Related Work}
\begin{figure*}[!t]
\centering
\includegraphics[width=0.94\textwidth]{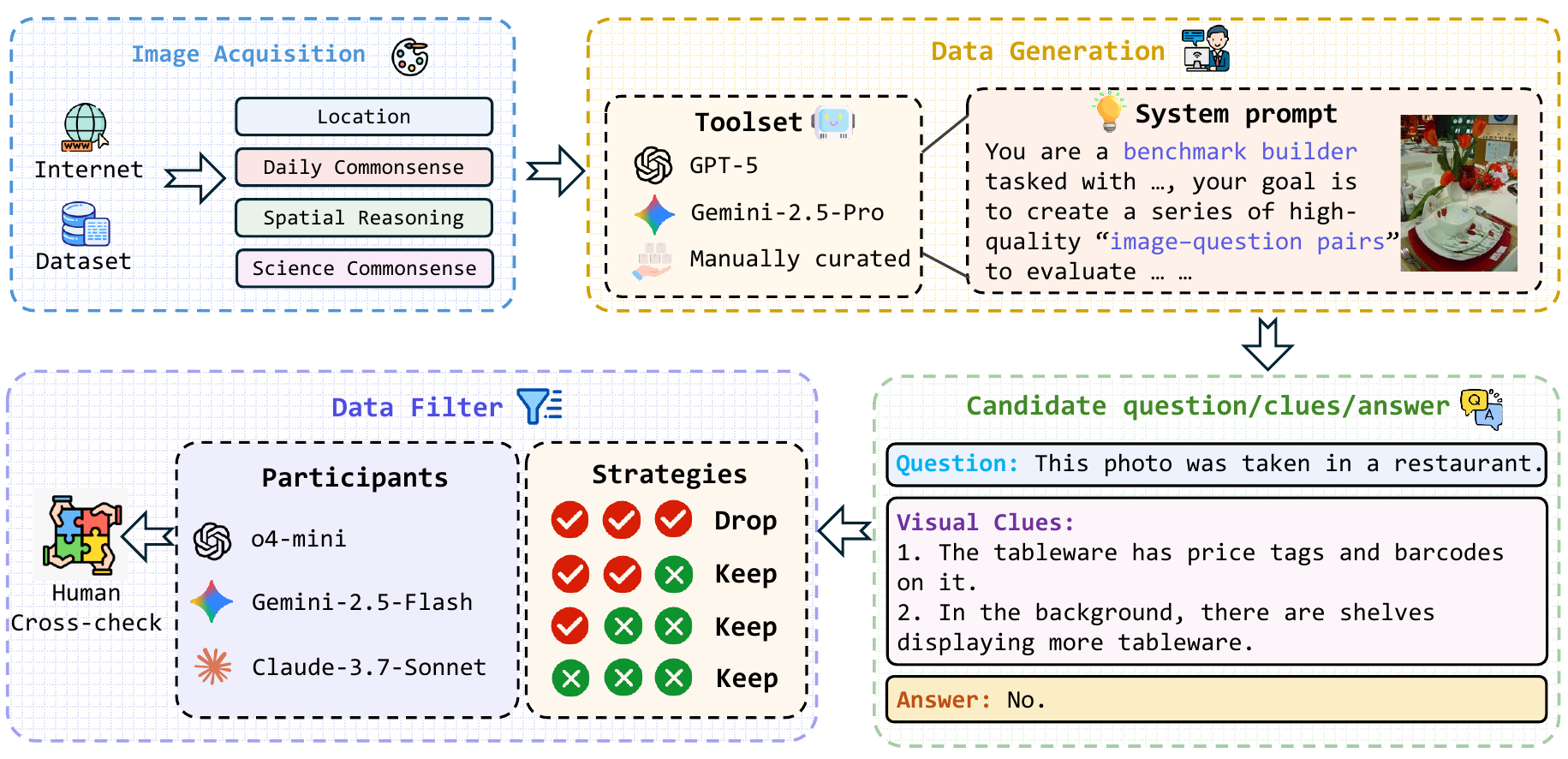}
\caption{Overview of the DailyClue construction pipeline. The process comprises three stages: (i) image collection across four categories; (ii) generation of question-clue-answer triplets; and (iii) data filtering.}
\label{fig:overall_framework}
\end{figure*}

\subsection{Benchmarks for MLLMs}
Existing MLLM benchmarks cover diverse categories, including general VQA~\cite{mmbench,realworldqa,simplevqa} as well as domain-specific question answering tasks such as physics reasoning~\cite{physbench,seephys}, mathematical problem solving~\cite{MathVision,mathverse,mathvista}, spatial reasoning~\cite{sibench,omnispatial}, color reasoning~\cite{colorbench}, and document understanding~\cite{ocrbench, docvqa}. 
Specifically, benchmarks centered on physics and mathematics primarily assess models’ mastery of discipline-specific knowledge, whereas those focused on color and spatial reasoning evaluate their ability to perceive fundamental object attributes and spatial relations. 
While works like OWLViz~\cite{owlviz} also explore daily scenes, DailyClue differs fundamentally: (1) Task Objective: instead of localizing object coordinates, we require models to deduce geographical contexts (e.g., country/city) from implicit visual evidence; (2) Reasoning Path: unlike OWLViz which may involve external tool-use, DailyClue is a self-contained testbed that isolates a model's ability to discover internal clues and leverage its own commonsense.
Collectively, benchmarks spanning these diverse domains have played an important role in diagnosing capability limitations of MLLMs from different perspectives.

\subsection{Benchmarks for Agentic Models} 
Following the release of o3, several think-with-image models~\cite{deepeyes,deepeyesv2,vacot,vlm-r3,pixel_reasoner,treebench,revpt,thyme,pyvision} and benchmarks~\cite{v_star,hrbench,seekworld,tir-bench} have emerged. As summarized in Table~\ref{tab:comparison_bench}, many existing benchmarks prioritize basic perceptual abilities~\cite{v_star,hrbench}, such as attribute recognition in high-resolution images, where targets require zooming to resolve.
Other benchmarks focus on vertical domains: SeekWorld~\cite{seekworld} targets the identification of the specific location; TIR-Bench~\cite{tir-bench} evaluates multi-disciplinary perception; and MME-RealWorld~\cite{mme-realworld}, despite utilizing real-world data, also incorporates some niche scenarios like remote sensing. While TreeBench~\cite{treebench} incorporates daily-life scenes, its questions remain simplistic, largely assessing basic perception.

Crucially, these benchmarks lack explicit visual clues, even though the pivotal role of such evidence is well-established~\cite{measuring,sherlock,visual-cot}. For instance, \cite{visual-cot} employs bounding box annotations to highlight the most informative regions, thereby training models to actively seek visual clues. Similarly, \cite{sherlock} demonstrates that systematically organizing visual clues facilitates more plausible inferences. To bridge this critical gap in the current landscape, we propose DailyClue.

\begin{table}[t]
\renewcommand\arraystretch{1.1}
\centering
\caption{
Comparison of benchmarks. DailyClue features explicit visual-clue annotations and emphasizes solving problems by leveraging these visual clues. 
}
\label{tab:comparison_bench}
\resizebox{0.48\textwidth}{!}{%
\fontsize{26pt}{30pt}\selectfont
\begin{tabular}{lccccc}
\toprule
\textbf{Benchmark} & \textbf{\makecell{Average \\ Resolution}} & \textbf{\makecell{Sample \\ Number}} & \textbf{\makecell{Task \\ Number}}  & \textbf{\makecell{Visual Clue \\ Annotation}} \\ 
\midrule
$V^*$ Bench & 2k & 191 & 2 & \textcolor{red}{\ding{55}} \\
HR-Bench 4K & 4k & 200 & 6 & \textcolor{red}{\ding{55}} \\
HR-Bench 8K & 8k & 200 & 6 & \textcolor{red}{\ding{55}} \\
SeekWorld & 1k & 693 & 1 & \textcolor{nvidiagreen}{\ding{51}} \\
TIR-Bench & 1k & 1215 & 13 & \textcolor{red}{\ding{55}} \\ 
MME-RealWorld & 2k & 29429 & 43 & \textcolor{red}{\ding{55}} \\
TreeBench & 2k & 405  & 10 & \textcolor{red}{\ding{55}} \\
\midrule
DailyClue & 1k & 666 & 16 & \textcolor{nvidiagreen}{\ding{51}} \\
\bottomrule
\end{tabular}
}
\end{table}

\section{DailyClue}
We introduce DailyClue, a benchmark anchored in visually rich and noise-intensive daily scenarios. We define ``daily activities'' as common behaviors performed by the general public outside of specialized professional domains (e.g., coding), primarily encompassing aspects such as clothing, food, housing, transportation, and social entertainment. The benchmark is designed to rigorously evaluate MLLMs' capability to locate decisive visual clues for reasoning across four core domains: location identification, spatial relationship reasoning, daily commonsense reasoning, and scientific commonsense reasoning.
In this section, we first detail the data collection pipeline~(\S\ref{sec:data_curation}) and then present a comprehensive statistical overview~(\S\ref{sec:benchmark_statistics}).

\subsection{Data Curation}
\label{sec:data_curation}

\paragraph{Data Collection.} 
We start by collecting images from relevant datasets for each category, prioritizing images with rich scene content to ensure informative visual clues. We supplement these with manually curated web samples; notably, the Daily and Scientific Commonsense subsets are predominantly sourced from the open web.
This manual curation is particularly necessary for the scientific domain, as existing benchmarks often focus on textbook-style problems~(e.g., circuit diagrams)~\cite{seephys,ir3d} or simple static states~\cite{treebench}, rather than scientific phenomena in daily-life scenarios. To bridge this gap, we construct the scientific subset using realistic imagery, incorporating both single-image scenes and, where necessary, multi-frame sequences to capture temporal dynamics.

\begin{figure}[t]
    \centering
    \includegraphics[width=0.48\textwidth]{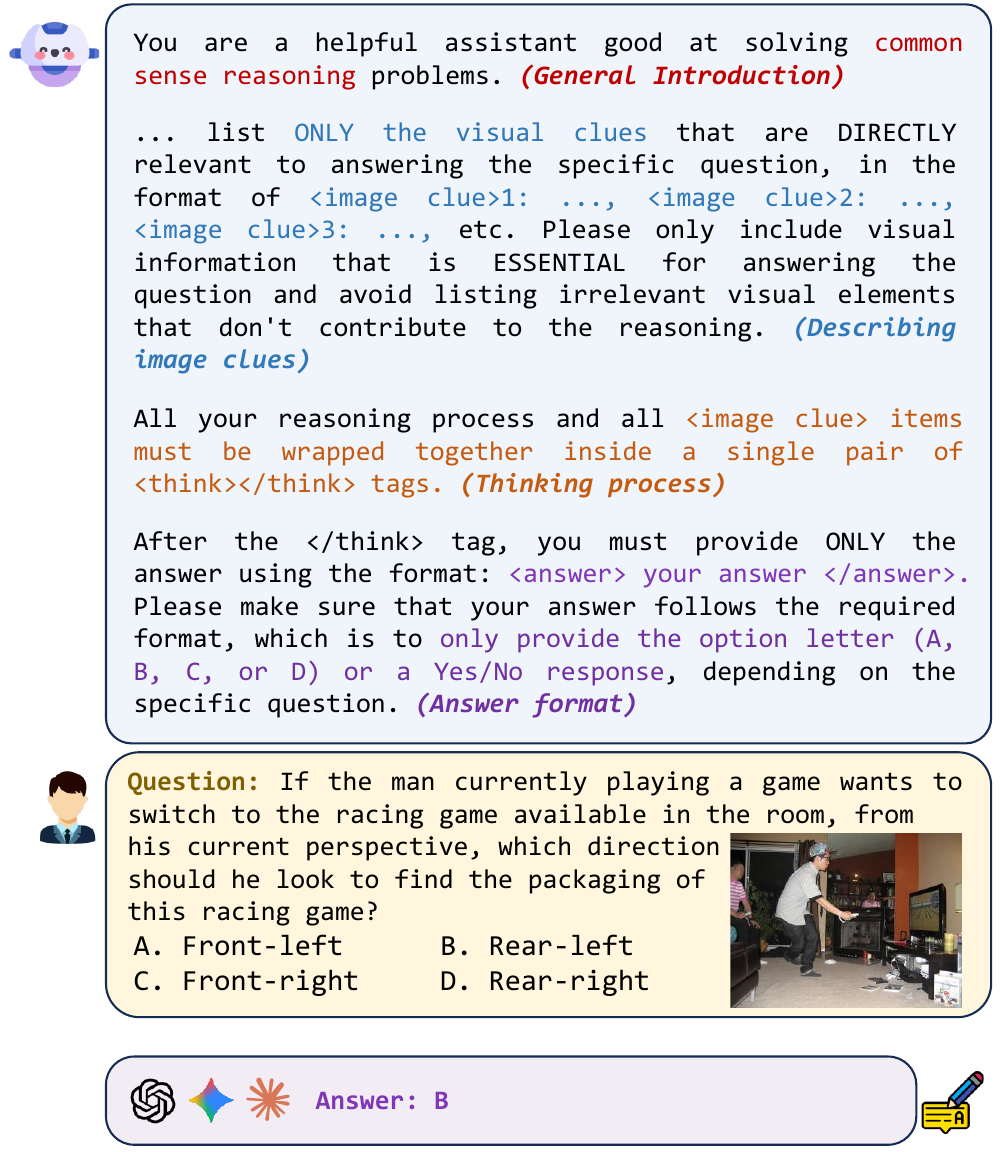}
    \caption{System and user prompts for the spatial relation reasoning task. In the user prompt, a question is posed to three MLLMs, and their responses are used to decide whether the corresponding triplet is retained.}
    \label{fig:filter_prompt}
\end{figure}

\begin{figure*}[!t]
\centering
\includegraphics[width=0.98\textwidth]{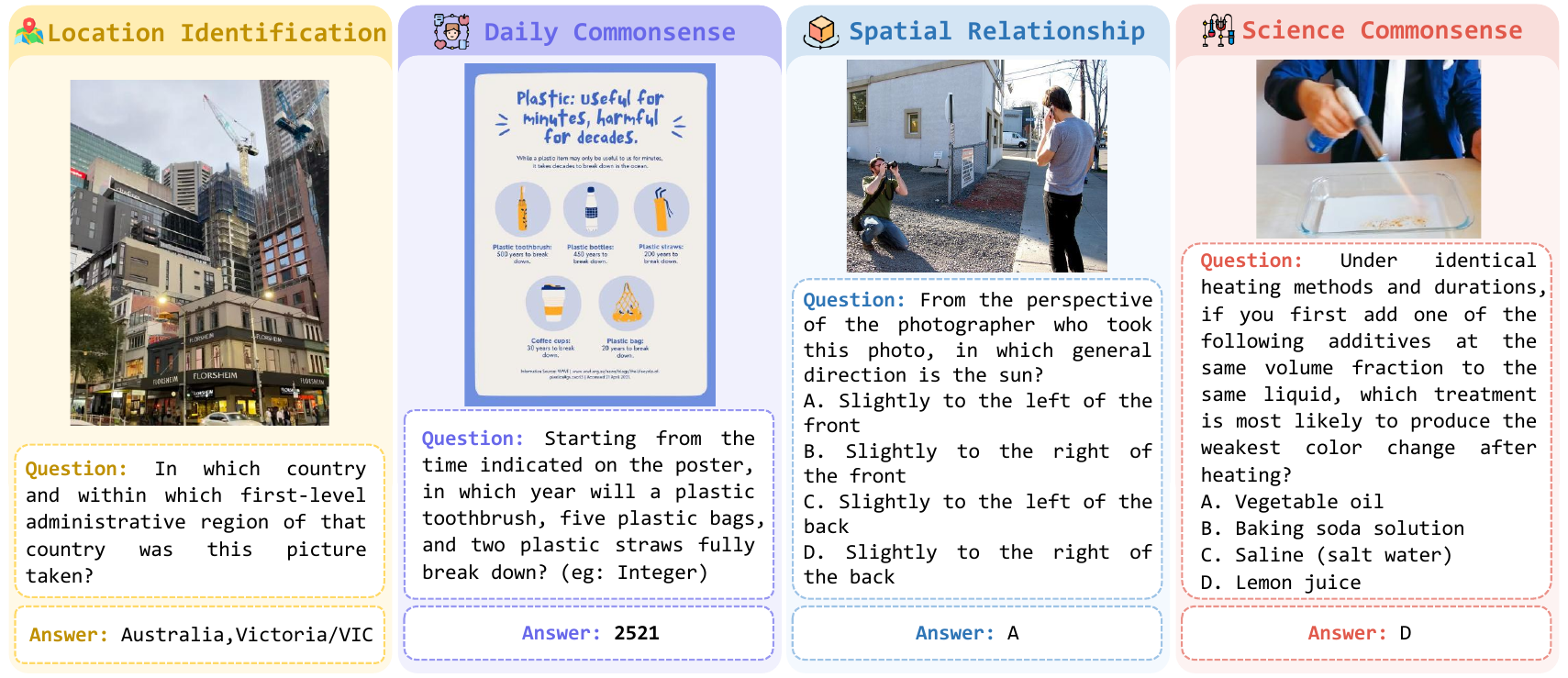}
\caption{Overview of DailyClue examples. DailyClue features four daily life scenarios across 16 reasoning subtasks. While some questions may appear trivial to humans, they pose significant challenges for MLLMs.}
\label{fig:example_viz}
\end{figure*}

\paragraph{Question–Clue–Answer Triplet Construction.}
We employ GPT-5~\cite{gpt5} and Gemini-2.5-Pro~\cite{gemini} to generate initial question–clue–answer triplets for each image. Given potential hallucinations, every triplet undergoes rigorous manual inspection. We strictly retain only challenging samples where the answer cannot be inferred from textual priors alone, but necessitates deep reasoning over visual clues. If any part of the generated triplet is flawed, we manually revise it or author a new one from scratch.
The design of system prompts is critical and strictly follows two core principles: (1) questions must be inherently indirect, avoiding explicit hints to compel independent visual evidence seeking; and (2) answers must be unambiguous and unique once the clue is identified. Prompts are further tailored to specific tasks; for instance, in spatial reasoning, the model is instructed to target positional relationships and occlusions. Detailed prompts are provided in Appendix~\ref{sec:appendix_system_prompt}.

\paragraph{Triplet Filtering and Verification.}
To ensure our benchmark possesses sufficient discriminative power to distinguish between MLLM capabilities, we target a high difficulty threshold. Accordingly, we implement a rigorous filtering pipeline for the generated triplets. GPT-o4-mini~\cite{gpt-4o}, Gemini-2.5-Flash~\cite{gemini}, and Claude-3.7-Sonnet are used as the automatic data filters.
By involving three distinct model families, we implement a consensus-based rejection strategy to mitigate model-specific quirks: if all three models answer a question correctly, the item is discarded as trivial; conversely, items answered correctly by at most two models are deemed sufficiently challenging and retained. The prompts used for this filtering process are illustrated in Figure~\ref{fig:filter_prompt}. Crucially, to prevent the benchmark from inheriting potential model-specific biases or artifacts, we do not rely solely on this automated process. Three annotators independently cross-check the correctness of each question, clue, and answer, ensuring that tasks remain strictly grounded in authentic, everyday logic rather than quirky model behaviors. This human-in-the-loop oversight ensures that the benchmark reflects human-defined difficulty and maintains high data rigor.

\subsection{Benchmark Overview}
\label{sec:benchmark_statistics}
\paragraph{Overall Task Definition.}
Drawing from common real-world scenarios—such as social interactions, transportation, and work—we categorize the benchmark into four primary types (Figure~\ref{fig:statistics}). 
Location Identification focuses on pinpointing the specific scene location. 
Spatial Reasoning evaluates the understanding of motion, occlusion, and spatial relationships. 
Daily Commonsense encompasses essential life aspects including food, health, and social customs. 
Finally, Scientific Commonsense necessitates applying domain knowledge in physics, chemistry, or biology to reason about scientific principles manifest in daily life.

\paragraph{Distribution of Each Subtask.}
The benchmark comprises 666 question-image pairs. Reflecting our emphasis on visual reasoning within daily-life contexts, we place significant weight on Location Identification, Daily Commonsense, and Spatial Reasoning. In contrast, Scientific Commonsense Reasoning, which necessitates specialized domain knowledge, accounts for a curated 18\% of the total. 
While all tasks are grounded in daily scenarios, the former categories pertain to commonplace situations, whereas the latter introduces a degree of technical specialization. Among these, Daily Commonsense Reasoning encompasses the most diverse array of subtasks, reflecting the multifaceted nature of real-life situations. Notably, the Planning and Consumption subtask comprises approximately 29\%~(53 questions) of this category. Reasoning about time and financial planning demands higher-order cognitive skills, which aligns precisely with the core objective of our benchmark.
Figure~\ref{fig:example_viz} presents representative examples, and additional examples are provided in Appendix~\ref{sec:more_expr_results}.
While the overall scale is relatively compact, this is a trade-off necessitated by our rigorous human-in-the-loop construction process. Each triplet undergoes meticulous manual verification to ensure high quality and reasoning depth, which is inherently more challenging than automated collection. Despite its size, our extensive evaluation of 25 MLLMs (see Table~\ref{tab:main_result}) demonstrates that the benchmark’s high difficulty and diversity across 16 sub-domains provide sufficient discriminative power to effectively assess the clue-finding and reasoning capabilities of MLLMs.

\paragraph{Data Sources and Question Types.}
The data sources and question types for each domain are shown in the right panel of Figure~\ref{fig:statistics}. Data for Location Identification and Daily Commonsense Reasoning are derived from a combination of existing datasets~\cite{ocrbench,seekworld,cosim} and curated Internet samples. In contrast, Spatial Reasoning is drawn exclusively from established benchmarks, while the Scientific Commonsense subset is entirely manually collected from the web. Collectively, the benchmark encompasses three distinct question formats.

\section{Experiment}
\begin{table*}[ht]
    \centering
    \begin{adjustbox}{width=0.94\textwidth}
    \small
    \begin{tabular}{l|ccccc}
    \toprule
        \textbf{Model} & \textbf{Overall} & \textbf{\makecell{Location \\ Identification}} & \textbf{\makecell{Spatial \\ Relationship}} & \textbf{\makecell{Daily  \\ Commonsense}} & \textbf{\makecell{Scientific \\ Commonsense}}  \\
        \midrule
        \rowcolor{background-grey}
        \multicolumn{6}{c}{\textbf{Open-source MLLMs}} \\
        LLaVA-OneVision-7B  & 24.47 & 10.50 & 34.97 & 25.56 & 31.71 \\
        LLaVA-OneVision-72B  & 33.18 & 15.50 & 47.85 & 33.33 & 42.28 \\
        LLaVA-OneVision-1.5-8B-Instruct & 29.43 & 10.50 & 47.85 & 27.78 & 38.21 \\
        InternVL3-8B  & 31.08 & 13.50 & 31.67 & 31.67 & 41.46 \\
        InternVL3-38B  & 36.94 & 17.00 & 47.85 & 47.22 & 39.84 \\
        InternVL3-78B  & 40.84 & 18.00 & \textbf{54.60} & \underline{52.78} & 42.28 \\
        InternVL-3.5-38B & 36.91 & 14.00 & \underline{49.69} & 43.33 & 43.90 \\
        Qwen2.5-VL-7B & 30.63 & 15.00 & 39.88 & 37.22 & 34.15 \\
        Qwen2.5-VL-32B & 35.59 & 21.50 & 42.94 & 42.78 & 38.21 \\
        Qwen2.5-VL-72B & \underline{40.84} & \textbf{24.50} & 47.85 & 48.33 & 47.15 \\
        Qwen3-VL-235B-A22B-Thinking  & \textbf{44.59} & \underline{23.00} & 49.08 & \textbf{56.67} & \textbf{56.10} \\
        Qwen3-VL-235B-A22B-Instruct  & 40.69 & 22.50 & 46.63 & 50.00 & \underline{48.78} \\
        \midrule
        \rowcolor{background-grey}
        \multicolumn{6}{c}{\textbf{Close-source MLLMs}} \\
        \raisebox{-0.3em}{\includegraphics[height=1.2em]{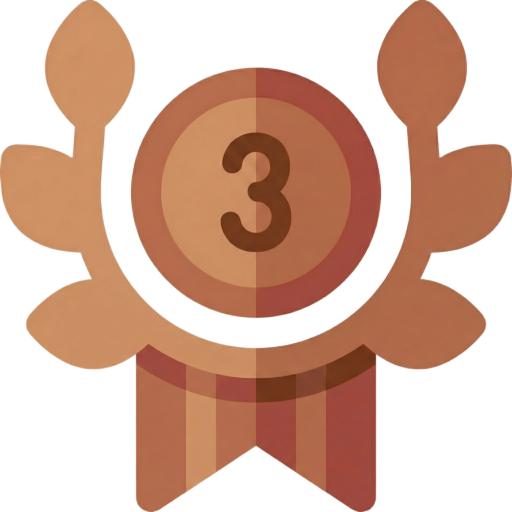}}
        Gemini-2.5-Flash  & 50.00 & 32.50 & 55.83 & \underline{59.44} & 56.91 \\
        \raisebox{-0.3em}{\includegraphics[height=1.2em]{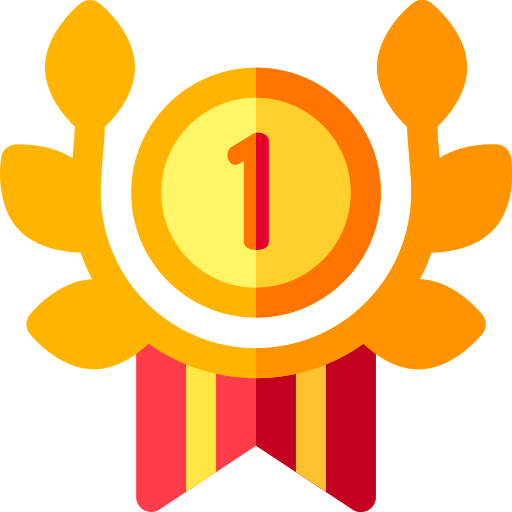}}
        Gemini-2.5-Pro  & \textbf{56.90} & \textbf{41.50} & \textbf{61.35} & \textbf{62.77} & \textbf{67.48} \\
        Claude-3.7-Sonnet & 41.14 & 18.50 & 57.06 & 47.22 & 47.97 \\
        Claude-sonnet-4  & 41.74 & 22.00 & 52.15 & 48.89 & 49.59 \\
        Claude-sonnet-4.5  & 41.74 & 21.00 & 53.99 & 49.44 & 47.97 \\
        o4-mini  & 47.00 & 25.50 & \underline{58.28} & 58.33 & 50.41 \\
        \raisebox{-0.3em}{\includegraphics[height=1.2em]{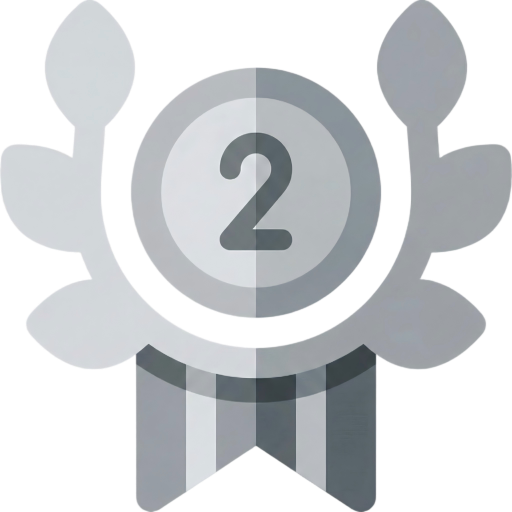}}
        GPT-5  & \underline{50.90} & \underline{38.00} & 57.67 & 51.67 & \underline{61.79} \\        
         \midrule
         \rowcolor{background-grey}
         \multicolumn{6}{c}{\textbf{Agentic Models}} \\
        DeepEyes-7B  & 30.93 & 18.50 & \underline{44.17} & 30.00 & 34.96 \\
        VLM-R3  & 33.18 & \underline{19.00} & 42.33 & \underline{36.11} & \underline{39.84} \\
        TreeVGR-7B  & 27.78 & 14.00 & 40.49 & 27.18 & 33.33 \\
        REVPT  & 25.83 & 6.50 & 38.04 & 32.22 & 31.71 \\
        Thyme  & \textbf{46.25} & \textbf{69.00} & 42.33 & 29.44 & 39.02 \\
        PyVision  & \underline{39.48} & 18.50 & \textbf{47.23} & \textbf{48.33} & \textbf{50.40} \\
        \midrule
        \rowcolor{background-grey}
        \multicolumn{6}{c}{\textbf{Human Baseline}} \\
        Human Baseline & 45.50 & 19.33 & 70.67 & 40.00 & 52.00 \\
    \bottomrule
    \end{tabular}
    \end{adjustbox}
    \caption{
    Performance comparison on the DailyClue benchmark. Metrics denote accuracy (\%). The best result is highlighted in \textbf{bold}, and the second best is \underline{underlined}.
    }
    \vspace{-0.3cm} 
    \label{tab:main_result} 
\end{table*}

\begin{table*}[ht]
    \centering
    \begin{adjustbox}{width=0.84\textwidth}
    \small
    \begin{tabular}{llllll}
    \toprule
        \multirow{2}{*}{\textbf{Model}} & \multicolumn{5}{c}{\textbf{Clue Source}} \\
        \cmidrule(lr){2-6}
        
         & \textbf{Qwen2.5-VL-72B} & \textbf{Claude-3.7} & \textbf{Gemini-2.5-Pro} & \textbf{GT Clue} & \textbf{No Clue}  \\
    \midrule
        
        Qwen2.5-VL-72B  & 40.09\down{-0.75} & 41.29\up{0.45} & 48.80\up{7.96} & \textbf{51.50}\up{10.66} & 40.84 \\
        
        Claude-3.7      & 42.49\up{1.35} & 43.39\up{2.25} & 51.95\up{10.81} & \textbf{56.00}\up{14.86} & 41.14 \\
        
        Gemini-2.5-Pro  & 52.85\down{-4.05} & 53.15\down{-3.75} & 55.26\down{-1.64} & \textbf{58.55}\up{1.65} & 56.90 \\    
    
    \bottomrule
    \end{tabular}
    \end{adjustbox}
    \caption{
    Effect of clue sources on MLLM reasoning. Columns denote the source of the clue provided to the target model (row). 
    ``No Clue'' denotes the baseline using intrinsic CoT without external context. Note that diagonal entries reflect performance when models are conditioned on their own explicitly self-generated clues.
    }
    \label{tab:exchange_clue} 
\end{table*}

\subsection{Experiment Setup}
\label{sec: expr}
\paragraph{MLLMs and Agentic Models.}
We categorize the evaluated models into three distinct groups: open-source models, close-source models, and tool-use capable agents. The open-source group comprises representative families such as the LLaVA series (e.g., LLaVA-OneVision-7B/72B~\cite{llava-ov}, LLaVA-OneVision-1.5~\cite{llava-ov-1.5}), the InternVL series (e.g., InternVL3-8B/38B/78B~\cite{internvl3}, InternVL3.5-38B~\cite{internvl3.5}), and the Qwen series (e.g., Qwen2.5-VL-7B/32B/72B~\cite{qwen2.5-vl}, Qwen3-VL-235B-A22B~\cite{qwen3-vl}). The close-source models encompass Gemini-2.5-Flash/Pro~\cite{gemini}, Claude-3.7/4~\cite{claude3}, and the GPT series (specifically GPT-5~\cite{gpt5} and o4-mini). Finally, the agentic models consist of DeepEyes~\cite{deepeyes}, VLM-R3~\cite{vlm-r3}, TreeVGR~\cite{treebench}, REVPT~\cite{revpt}, Thyme~\cite{thyme}, and PyVision~\cite{pyvision}.

\paragraph{Human Baseline.}
To establish a human baseline for the understanding of daily-life scenarios, we conduct a rigorous user study. Specifically, we design four task-specific questionnaires corresponding to the benchmark's primary domains, each featuring 50 representative, randomly selected samples. We recruit six undergraduate participants to answer these questions independently, establishing a reference standard for assessing MLLMs' performance. Further details regarding the questionnaires are provided in Appendix~\ref{sec:human_baseline}.

\subsection{Evaluation}
\label{sec:eval}
\paragraph{General Evaluation Protocol.}
For the Location Identification task, we adopt the evaluation protocol aligned with
~\cite{seekworld}. Specifically, we enforce strict exact matching for country-level predictions. Conversely, for first-level administrative divisions, we apply a relaxed criterion: the ground truth is treated as a candidate set encompassing official names, abbreviations, and aliases. A prediction is deemed correct if it matches any entry within this set.
For other tasks with multiple-choice and Yes/No questions, we evaluate their performance using exact string matching. For open-ended questions, we employ Gemini-2.5-Pro as the automated judge model.

\paragraph{Rigorous Evaluation Protocol.}
To mitigate the risk of spurious correctness—where models arrive at the correct answer through ``\textit{lucky guesses}''—we introduce a rigorous dual-verification mechanism. Beyond merely verifying answer accuracy, we enforce a more rigorous constraint requiring that the visual clues identified by the model explicitly intersect with the ground truth.
Formally, given a dataset $\mathcal{D} = \{(Q_i, A_i, c_i)\}_{i=1}^N$, let $\hat{A}_i$ be the predicted answer and $\hat{c}_i$ denote the visual clues extracted from the model's CoT process.
The strict accuracy score $S(i)$ is defined as:
\begin{equation}
    S(i) = 
    \begin{cases} 
      1, & \text{if } \hat{A}_i = A_i \quad \text{and} \quad \hat{c}_i \cap c_i \neq \emptyset \\
      0, & \text{otherwise}
    \end{cases}
\end{equation}

\subsection{Main Results}
Table~\ref{tab:main_result} presents a comprehensive benchmark of open-source, proprietary, and agentic models across four tasks. In general, proprietary models dominate the leaderboard, with Gemini-2.5-Pro establishing the state-of-the-art performance. Other key observations are summarized below:

\paragraph{Finding 1: DailyClue poses a significant challenge to existing MLLMs. }
Notably, models struggle to exceed 60\% accuracy. Among open-source models, Qwen3-VL-235B-A22B-Thinking emerges as the top open-source model (44.59\%), whereas the proprietary Gemini-2.5-Pro establishes the state-of-the-art at 56.90\%. 
While Gemini Pro leads the leaderboard, its performance remains far from saturation (typically >90\%), leaving substantial room for improvement. This observation aligns with other high-quality reasoning benchmarks, such as TreeBench~\cite{treebench} and MME-RealWorld~\cite{mme-realworld}, where top-tier models consistently fall within the 50\%–60\% range. This consistency demonstrates that DailyClue accurately captures the current frontiers of multimodal reasoning rather than introducing idiosyncratic difficulty.
The performance gap largely stems from the superior capacity of these top-tier models to extract precise visual clues and engage in rigorous logical reasoning. Additionally, within specific model families, scaling laws remain evident: performance correlates positively with model size. Models with comparable parameter counts (e.g., InternVL3-38B vs. Qwen2.5-VL-32B) exhibit highly similar accuracy levels.

\paragraph{Finding 2: Agentic models lag behind proprietary counterparts despite their integration of external tool.} 
Most evaluated agents are fine-tuned on Qwen2.5-VL-7B yet show only negligible gains on our benchmark. Thyme emerges as the top performer, specifically dominating Location Identification (outperforming Gemini-2.5-Pro by 27.5\%). However, it remains uncompetitive across other tasks. This specialization is likely due to its training data distribution, which covers street signs and salient landmarks. These specific examples likely enable the model to better identify key visual clues essential for geolocation tasks.

\paragraph{Finding 3: MLLMs surpass humans in knowledge-intensive and computational tasks, while humans retain an edge in spatial intuition.}
Although average human accuracy trails Gemini-2.5-Pro by 11.40\%, the performance gap varies substantially across different domains. Notably, humans demonstrate superior capability in Spatial Reasoning. The lower overall human performance is primarily driven by Location Identification, as annotators often lack the encyclopedic knowledge required to recognize obscure locales. 

\section{Further Findings}
\subsection{Probing MLLMs with Explicit Clues}
\begin{tcolorbox}[colframe=black,
arc=2pt,
boxsep=-0.35em,
left=8pt,right=8pt,
]
    \paragraph{\textbf{{Insight} 1.}} Despite the strong reasoning capabilities of current MLLMs, their performance is heavily bottlenecked by the failure to accurately capture critical visual semantics. Enhancing visual clues leads to substantial accuracy improvements across models of varying scales.
\end{tcolorbox}
\noindent
To investigate the influence of visual clues, we conditioned Qwen2.5-VL-72B, Claude-3.7, and Gemini-2.5-Pro on clues from diverse sources.
As shown in Table~\ref{tab:exchange_clue}, performance exhibits a strict monotonic increase corresponding to clue quality (Qwen < Claude < Gemini < GT). 
Notably, GT clue injection yields substantial gains across all models (e.g., Claude +14.86\%), confirming that while current reasoning capabilities are robust, performance remains constrained by inaccurate clue extraction.

Moreover, Gemini-2.5-Pro emerges as both the superior solver and clue provider. However, its performance degrades significantly (dropping $\sim$4\%) when conditioned on inferior clues from Qwen or Claude. This suggests that misleading external clues can override the model's judgment and induce hallucinations, thereby disrupting its originally correct reasoning trajectory. We attribute this vulnerability to \textit{textual bias} and \textit{visual sycophancy}~\cite{zheng2025unveiling, pi2025pointing}: MLLMs tend to prioritize textual context over their own visual perception, causing them to align with misleading clues rather than correcting them.

\begin{figure}[t]
    \centering
    \hspace{-0.3cm} 
    \includegraphics[width=0.46\textwidth]{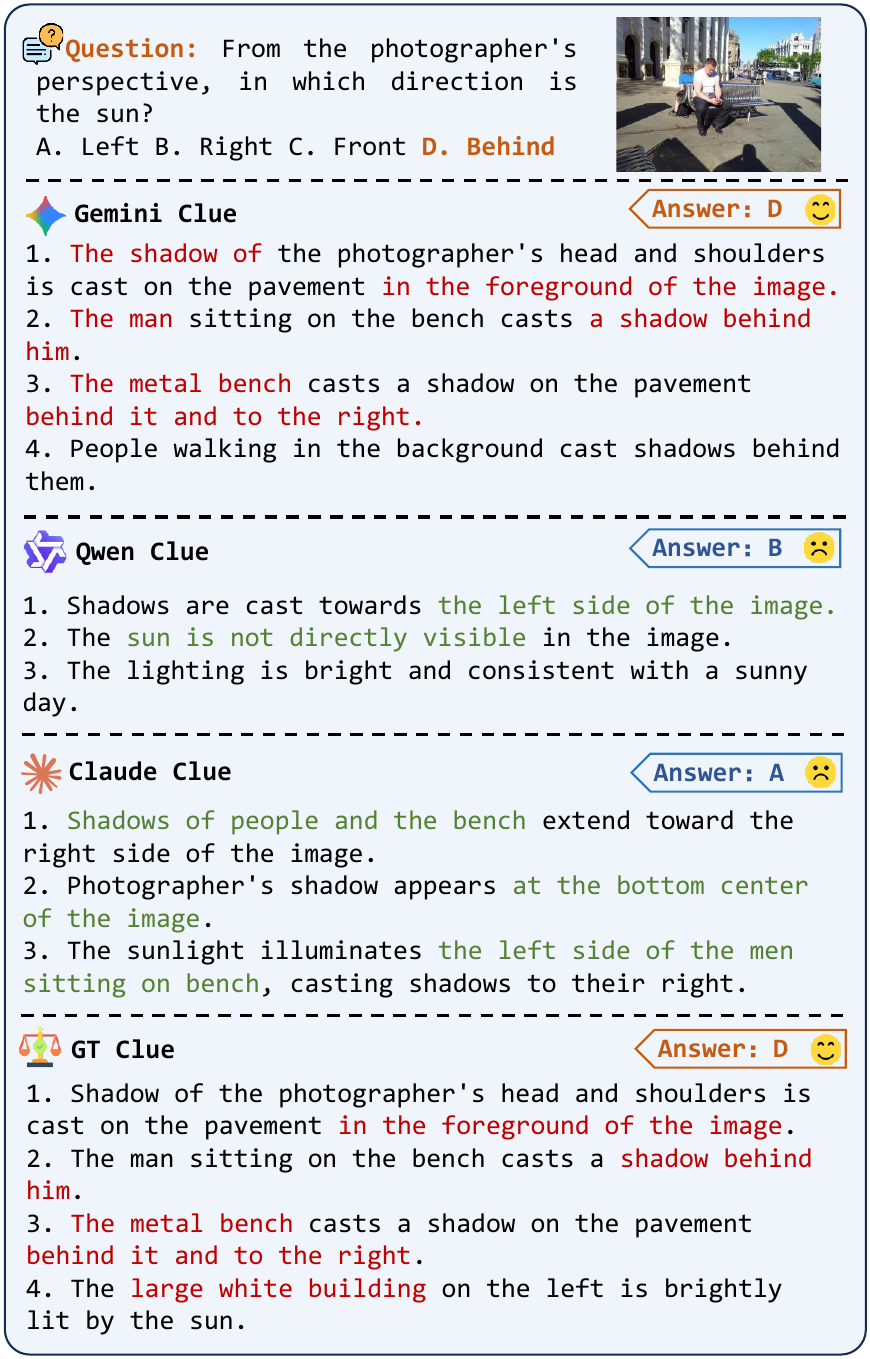}
    \caption{Comparison of answer generation under different clue contexts. We feed Claude-3.7 with visual clues from varying sources as additional context. As observed, clues generated by MLLMs may be useless~(\textcolor{text-green}{green}), causing the solver to fail. Conversely, the GT clue provides precise visual grounding (\textcolor{text-red}{red}), acting as the only effective guide for correct reasoning.}
    \label{fig:exchange_clue}
    \vspace{-0.3cm}
\end{figure}

\begin{figure}[t]
    \centering
    \includegraphics[width=0.95\linewidth]{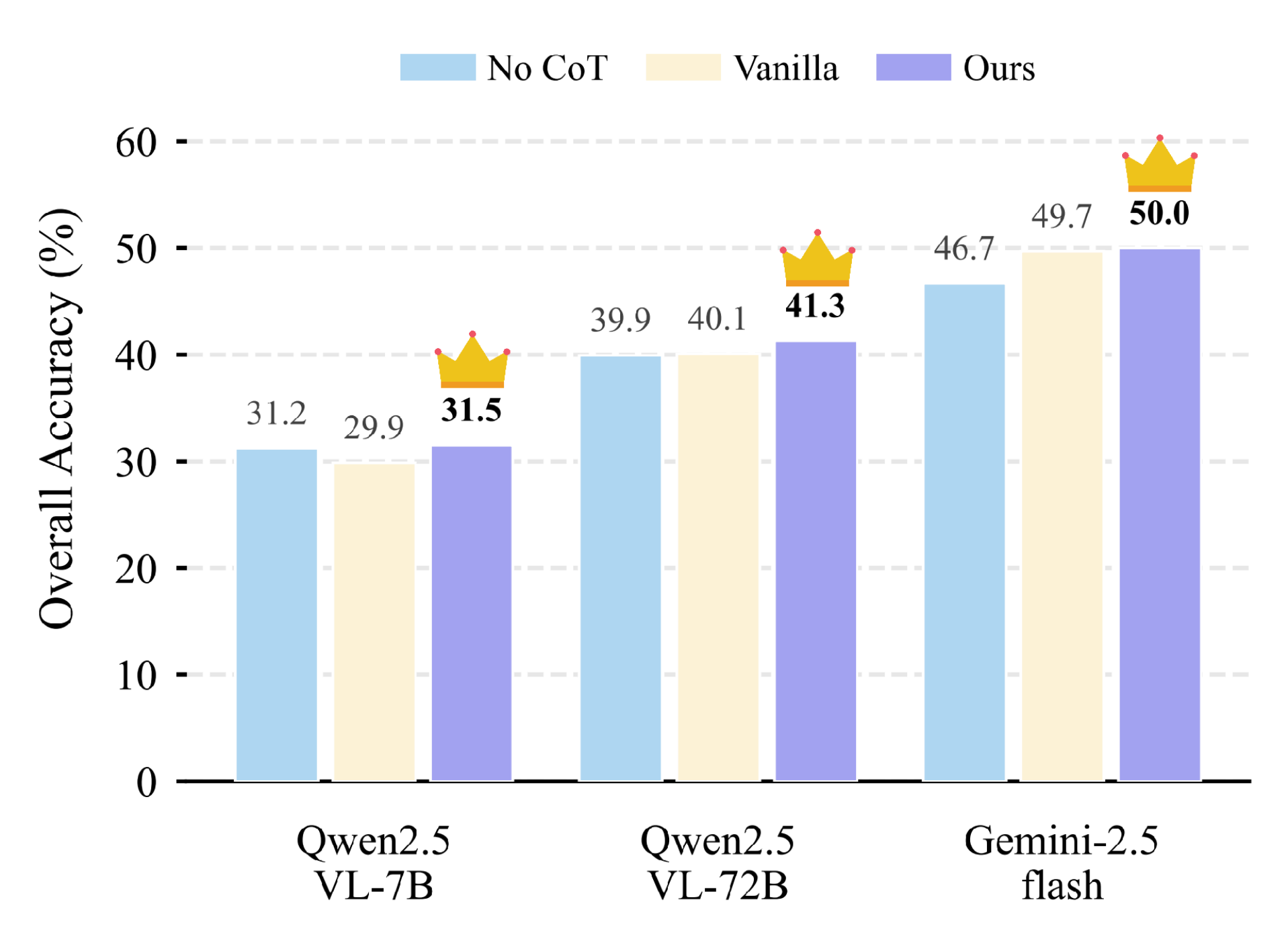}
    \caption{\textbf{Impact of visual-clue-reasoning on accuracy.} Our Clue-guided CoT (Purple) consistently outperforms baselines across all models.}
    \label{fig:ablation_clue}
\end{figure}

\begin{figure}[t]
    \centering
    \includegraphics[width=0.95\linewidth]{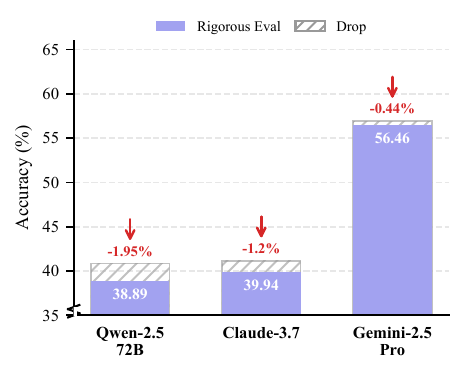}
    \caption{\textbf{Accuracy comparison between General and Rigorous Evaluation Protocols.} The purple region denotes the Rigorous accuracy, whereas the full bar height (including the gray ‘Drop’ area) corresponds to the General accuracy.}
    \label{fig:rigorous_evaluation}
\end{figure}

To intuitively demonstrate the impact of external visual clues, Figure~\ref{fig:exchange_clue} illustrates Claude-3.7's reasoning when conditioned on clues from different sources. We observe that Claude-3.7 successfully derives the correct answer using a clue from Gemini, but fails when relying on the weaker model, Qwen. The specific user prompts used for clue injection are detailed in the Appendix~\ref{sec:appendix_system_prompt}.

\subsection{Efficacy of Visual Clue-Driven Reasoning}
\begin{tcolorbox}[colframe=black,
arc=2pt,
boxsep=-0.35em,
left=8pt,right=8pt,
]
    \paragraph{\textbf{{Insight} 2.}} Mandating active visual clues within CoT acts as a critical anchor for the reasoning process. This constraint effectively mitigates reasoning drift, thereby significantly improving the model's final accuracy.
\end{tcolorbox}

\noindent
To validate the efficacy of explicitly requiring MLLMs to attend to visual clues during the CoT process, we conduct a comprehensive ablation study. Specifically, we evaluate the models under three distinct inference settings: (1) direct answering without reasoning (No CoT), (2) reasoning with default thinking process (Vanilla CoT), and (3) actively identifying visual clues during the reasoning process (Ours). Detailed system prompts are provided in Appendix~\ref{sec:appendix_system_prompt}.

As illustrated in Figure~\ref{fig:ablation_clue}, all three models exhibit a consistent upward trend in accuracy across these settings. Notably, actively engaging with visual clues during the reasoning phase significantly facilitates correct predictions~\cite{vlm2-bench}. By explicitly prompting the model to utilize visual clues, we impose a constraint on the CoT, which effectively grounds the reasoning process and mitigates reasoning drift.

\subsection{Rigorous Evaluation with Visual Clues}

\begin{tcolorbox}[colframe=black,
arc=2pt,
boxsep=-0.35em,
left=8pt,right=8pt,
]
    \paragraph{\textbf{{Insight} 3.}} 
    Superficial accuracy may belie a critical disconnect between reasoning and prediction. We reveal that correct answers sometimes stem from invalid logic—specifically \textit{illusionary} and \textit{useless clues}.
\end{tcolorbox}

\noindent
To illustrate the ``right answer, wrong reason'' phenomenon, we apply the rigorous evaluation protocol described in Section~\ref{sec:eval} to examine the intermediate visual clues. We identify two main types of invalid reasoning: \textit{illusionary clues}~(hallucinating non-existent objects) and \textit{useless clues}~(using correct but irrelevant objects).

As shown in Figure~\ref{fig:rigorous_evaluation}, the overall performance drops are relatively small, suggesting that top models possess a strong capability in locating relevant visual clues, and their reasoning is generally grounded rather than random guessing.
However, Qwen2.5-VL-72B and Claude-3.7 still show noticeable accuracy drops of 1.95\% and 1.2\%, indicating they sometimes guess correctly based on wrong clues. In contrast, Gemini-2.5-Pro is exceptionally stable with a negligible drop of only 0.44\%, demonstrating that its reasoning is highly consistent with its predictions, achieving superior reasoning fidelity.

\section{Conclusion}
In this paper, we introduce DailyClue, a challenging benchmark designed to evaluate visual reasoning in daily-centric scenarios. Unlike existing benchmarks, DailyClue prioritizes authentic daily environments characterized by visual complexity, where filtering noise and identifying pertinent visual clues are essential for accurate reasoning. 
To ensure challenging and reliable data, we employ a collaborative pipeline involving top-tier models to synthesize initial question-clue-answer triplets, followed by strict multi-round filtering. Comprehensive evaluation across 25 MLLMs substantiates the pivotal role of visual clues, revealing that the accurate identification of visual evidence serves as the core factor in achieving correct predictions.

\section*{Limitations}

DailyClue focuses on four major daily-life domains and sixteen sub-tasks. While representative, these settings do not exhaustively cover the full diversity of real-world scenarios. The current scale remains relatively compact due to our rigorous human-in-the-loop verification, which prioritizes reasoning depth over raw volume. We plan to address this by incorporating a broader array of scenarios and expanding the data volume in future versions.
Moreover, our evaluation excludes certain high-cost proprietary models (e.g., Gemini-3-Pro) due to practical constraints, which may limit the breadth of model comparisons.
Finally, the current version of DailyClue emphasizes static visual reasoning. While multimodal research is increasingly exploring video-based and interactive settings, they remain beyond the current scope and are reserved for future investigation.
\section*{Ethical Considerations}
Our research focuses on evaluating MLLMs within complex, real-world scenarios through a benchmark synthesized from established public datasets and meticulously curated natural scenes. 
The dataset is strictly for academic research. We adhere to the licenses of all open-source data used. Our annotation workflow is designed to prioritize privacy. Annotators are tasked exclusively with quality assurance, thereby mitigating any potential risk of handling or compromising sensitive personal data.

\section*{Acknowledgments}
The paper is supported in part by the National Natural Science Foundation of China under grant Nos.62441231, 62293542, 62472065, U23B2010, Liao Ning Province Science and Technology Plan No.2023JH26/10200016, Dalian City Science and Technology Innovation Fund No. 2023JJ11CG001 and Ningbo Key R\&D project under Grant No. 2025Z039.
    

\bibliography{main}

\begin{thebibliography}{51}
\providecommand{\natexlab}[1]{#1}

\bibitem[{An et~al.(2025)An, Xie, Yang, Zhang, Zhao, Cheng, Wang, Xu, Chen, Wu et~al.}]{llava-ov-1.5}
Xiang An, Yin Xie, Kaicheng Yang, Wenkang Zhang, Xiuwei Zhao, Zheng Cheng, Yirui Wang, Songcen Xu, Changrui Chen, Chunsheng Wu, and 1 others. 2025.
\newblock Llava-onevision-1.5: Fully open framework for democratized multimodal training.
\newblock \emph{arXiv preprint arXiv:2509.23661}.

\bibitem[{Anthropic(2024)}]{claude3}
Anthropic. 2024.
\newblock The claude 3 model family: Opus, sonnet, haiku.
\newblock \url{https://www-cdn.anthropic.com/de8ba9b01c9ab7cbabf5c33b80b7bbc618857627/Model_Card_Claude_3.pdf}.

\bibitem[{Bai et~al.(2025{\natexlab{a}})Bai, Cai, Chen, Chen, Chen, Cheng, Deng, Ding, Gao, Ge et~al.}]{qwen3-vl}
Shuai Bai, Yuxuan Cai, Ruizhe Chen, Keqin Chen, Xionghui Chen, Zesen Cheng, Lianghao Deng, Wei Ding, Chang Gao, Chunjiang Ge, and 1 others. 2025{\natexlab{a}}.
\newblock Qwen3-vl technical report.
\newblock \emph{arXiv preprint arXiv:2511.21631}.

\bibitem[{Bai et~al.(2025{\natexlab{b}})Bai, Chen, Liu, Wang, Ge, Song, Dang, Wang, Wang, Tang et~al.}]{qwen2.5-vl}
Shuai Bai, Keqin Chen, Xuejing Liu, Jialin Wang, Wenbin Ge, Sibo Song, Kai Dang, Peng Wang, Shijie Wang, Jun Tang, and 1 others. 2025{\natexlab{b}}.
\newblock Qwen2. 5-vl technical report.
\newblock \emph{arXiv preprint arXiv:2502.13923}.

\bibitem[{Chen et~al.(2024)Chen, Sikka, Cogswell, Ji, and Divakaran}]{measuring}
Yangyi Chen, Karan Sikka, Michael Cogswell, Heng Ji, and Ajay Divakaran. 2024.
\newblock Measuring and improving chain-of-thought reasoning in vision-language models.
\newblock In \emph{Proceedings of the 2024 Conference of the North American Chapter of the Association for Computational Linguistics: Human Language Technologies (Volume 1: Long Papers)}, pages 192--210.

\bibitem[{Cheng et~al.(2025)Cheng, Zhang, Zhang, Yang, Guan, Wu, Li, Zhang, Liu, Mai et~al.}]{simplevqa}
Xianfu Cheng, Wei Zhang, Shiwei Zhang, Jian Yang, Xiangyuan Guan, Xianjie Wu, Xiang Li, Ge~Zhang, Jiaheng Liu, Yuying Mai, and 1 others. 2025.
\newblock Simplevqa: Multimodal factuality evaluation for multimodal large language models.
\newblock In \emph{Proceedings of the IEEE/CVF International Conference on Computer Vision}, pages 4637--4646.

\bibitem[{Chow et~al.(2025)Chow, Mao, Li, Seita, Guizilini, and Wang}]{physbench}
Wei Chow, Jiageng Mao, Boyi Li, Daniel Seita, Vitor Guizilini, and Yue Wang. 2025.
\newblock Physbench: Benchmarking and enhancing vision-language models for physical world understanding.
\newblock \emph{arXiv preprint arXiv:2501.16411}.

\bibitem[{Comanici et~al.(2025)Comanici, Bieber, Schaekermann, Pasupat, Sachdeva, Dhillon, Blistein, Ram, Zhang, Rosen et~al.}]{gemini}
Gheorghe Comanici, Eric Bieber, Mike Schaekermann, Ice Pasupat, Noveen Sachdeva, Inderjit Dhillon, Marcel Blistein, Ori Ram, Dan Zhang, Evan Rosen, and 1 others. 2025.
\newblock Gemini 2.5: Pushing the frontier with advanced reasoning, multimodality, long context, and next generation agentic capabilities.
\newblock \emph{arXiv preprint arXiv:2507.06261}.

\bibitem[{Hessel et~al.(2022)Hessel, Hwang, Park, Zellers, Bhagavatula, Rohrbach, Saenko, and Choi}]{sherlock}
Jack Hessel, Jena~D Hwang, Jae~Sung Park, Rowan Zellers, Chandra Bhagavatula, Anna Rohrbach, Kate Saenko, and Yejin Choi. 2022.
\newblock The abduction of sherlock holmes: A dataset for visual abductive reasoning.
\newblock In \emph{European Conference on Computer Vision}, pages 558--575. Springer.

\bibitem[{Hong et~al.(2025)Hong, Zhao, Zhu, Lu, Xu, and Yu}]{deepeyesv2}
Jack Hong, Chenxiao Zhao, ChengLin Zhu, Weiheng Lu, Guohai Xu, and Xing Yu. 2025.
\newblock Deepeyesv2: Toward agentic multimodal model.
\newblock \emph{arXiv preprint arXiv:2511.05271}.

\bibitem[{Hurst et~al.(2024)Hurst, Lerer, Goucher, Perelman, Ramesh, Clark, Ostrow, Welihinda, Hayes, Radford et~al.}]{gpt-4o}
Aaron Hurst, Adam Lerer, Adam~P Goucher, Adam Perelman, Aditya Ramesh, Aidan Clark, AJ~Ostrow, Akila Welihinda, Alan Hayes, Alec Radford, and 1 others. 2024.
\newblock Gpt-4o system card.
\newblock \emph{arXiv preprint arXiv:2410.21276}.

\bibitem[{Jia et~al.(2025)Jia, Qi, Zhang, Zhang, Yu, He, Wang, and Yi}]{omnispatial}
Mengdi Jia, Zekun Qi, Shaochen Zhang, Wenyao Zhang, Xinqiang Yu, Jiawei He, He~Wang, and Li~Yi. 2025.
\newblock Omnispatial: Towards comprehensive spatial reasoning benchmark for vision language models.
\newblock \emph{arXiv preprint arXiv:2506.03135}.

\bibitem[{Jiang et~al.(2025)Jiang, Heng, Ye, Yang, Xu, Yan, Zhang, Huang, and Zhang}]{vlm-r3}
Chaoya Jiang, Yongrui Heng, Wei Ye, Han Yang, Haiyang Xu, Ming Yan, Ji~Zhang, Fei Huang, and Shikun Zhang. 2025.
\newblock Vlm-r$^3$: Region recognition, reasoning, and refinement for enhanced multimodal chain-of-thought.
\newblock \emph{arXiv preprint arXiv:2505.16192}.

\bibitem[{Kim et~al.(2022)Kim, Zala, and Bansal}]{cosim}
Hyounghun Kim, Abhay Zala, and Mohit Bansal. 2022.
\newblock Cosim: Commonsense reasoning for counterfactual scene imagination.
\newblock \emph{arXiv preprint arXiv:2207.03961}.

\bibitem[{Li et~al.(2024)Li, Zhang, Guo, Zhang, Li, Zhang, Zhang, Zhang, Li, Liu et~al.}]{llava-ov}
Bo~Li, Yuanhan Zhang, Dong Guo, Renrui Zhang, Feng Li, Hao Zhang, Kaichen Zhang, Peiyuan Zhang, Yanwei Li, Ziwei Liu, and 1 others. 2024.
\newblock Llava-onevision: Easy visual task transfer.
\newblock \emph{arXiv preprint arXiv:2408.03326}.

\bibitem[{Li et~al.(2025{\natexlab{a}})Li, Zhong, Zhao, Zhang, Lin, Lai, Chen, Psounis, and Zhang}]{tir-bench}
Ming Li, Jike Zhong, Shitian Zhao, Haoquan Zhang, Shaoheng Lin, Yuxiang Lai, Wei Chen, Konstantinos Psounis, and Kaipeng Zhang. 2025{\natexlab{a}}.
\newblock Tir-bench: A comprehensive benchmark for agentic thinking-with-images reasoning.
\newblock \emph{arXiv preprint arXiv:2511.01833}.

\bibitem[{Li et~al.(2025{\natexlab{b}})Li, Liu, Isobe, Jia, Cui, Zhou, Li, He, Lu, Wang et~al.}]{reneg}
Xiaomin Li, Yixuan Liu, Takashi Isobe, Xu~Jia, Qinpeng Cui, Dong Zhou, Dong Li, You He, Huchuan Lu, Zhongdao Wang, and 1 others. 2025{\natexlab{b}}.
\newblock Reneg: Learning negative embedding with reward guidance.
\newblock In \emph{Proceedings of the Computer Vision and Pattern Recognition Conference}, pages 23636--23645.

\bibitem[{Liang et~al.(2025)Liang, Li, Fan, Li, Nguyen, Cobbina, Bhardwaj, Chen, Liu, and Zhou}]{colorbench}
Yijun Liang, Ming Li, Chenrui Fan, Ziyue Li, Dang Nguyen, Kwesi Cobbina, Shweta Bhardwaj, Jiuhai Chen, Fuxiao Liu, and Tianyi Zhou. 2025.
\newblock Colorbench: Can vlms see and understand the colorful world? a comprehensive benchmark for color perception, reasoning, and robustness.
\newblock \emph{arXiv preprint arXiv:2504.10514}.

\bibitem[{Liu et~al.(2025)Liu, Li, Li, Wu, Li, Yang, Zhang, Lin, Han, and Feng}]{ir3d}
Parker Liu, Chenxin Li, Zhengxin Li, Yipeng Wu, Wuyang Li, Zhiqin Yang, Zhenyuan Zhang, Yunlong Lin, Sirui Han, and Brandon~Y Feng. 2025.
\newblock Ir3d-bench: Evaluating vision-language model scene understanding as agentic inverse rendering.
\newblock \emph{arXiv preprint arXiv:2506.23329}.

\bibitem[{Liu et~al.(2024{\natexlab{a}})Liu, Duan, Zhang, Li, Zhang, Zhao, Yuan, Wang, He, Liu et~al.}]{mmbench}
Yuan Liu, Haodong Duan, Yuanhan Zhang, Bo~Li, Songyang Zhang, Wangbo Zhao, Yike Yuan, Jiaqi Wang, Conghui He, Ziwei Liu, and 1 others. 2024{\natexlab{a}}.
\newblock Mmbench: Is your multi-modal model an all-around player?
\newblock In \emph{European conference on computer vision}, pages 216--233. Springer.

\bibitem[{Liu et~al.(2024{\natexlab{b}})Liu, Li, Huang, Yang, Yu, Li, Yin, Liu, Jin, and Bai}]{ocrbench}
Yuliang Liu, Zhang Li, Mingxin Huang, Biao Yang, Wenwen Yu, Chunyuan Li, Xu-Cheng Yin, Cheng-Lin Liu, Lianwen Jin, and Xiang Bai. 2024{\natexlab{b}}.
\newblock Ocrbench: on the hidden mystery of ocr in large multimodal models.
\newblock \emph{Science China Information Sciences}, 67(12):220102.

\bibitem[{Lu et~al.(2023)Lu, Bansal, Xia, Liu, Li, Hajishirzi, Cheng, Chang, Galley, and Gao}]{mathvista}
Pan Lu, Hritik Bansal, Tony Xia, Jiacheng Liu, Chunyuan Li, Hannaneh Hajishirzi, Hao Cheng, Kai-Wei Chang, Michel Galley, and Jianfeng Gao. 2023.
\newblock Mathvista: Evaluating mathematical reasoning of foundation models in visual contexts.
\newblock \emph{arXiv preprint arXiv:2310.02255}.

\bibitem[{Mathew et~al.(2021)Mathew, Karatzas, and Jawahar}]{docvqa}
Minesh Mathew, Dimosthenis Karatzas, and CV~Jawahar. 2021.
\newblock Docvqa: A dataset for vqa on document images.
\newblock In \emph{Proceedings of the IEEE/CVF winter conference on applications of computer vision}, pages 2200--2209.

\bibitem[{Nguyen et~al.(2025)Nguyen, Nguyen, Nguyen, Luong, Dang, and Lai}]{owlviz}
Thuy Nguyen, Dang Nguyen, Hoang Nguyen, Thuan Luong, Long~Hoang Dang, and Viet~Dac Lai. 2025.
\newblock Owlviz: An open-world benchmark for visual question answering.
\newblock \emph{arXiv preprint arXiv:2503.07631}.

\bibitem[{OpenAI(2025)}]{gpt5}
OpenAI. 2025.
\newblock Gpt-5.
\newblock \url{https://openai.com/research}.
\newblock Accessed: 2025-08-14.

\bibitem[{Peng et~al.()Peng, Wang, Tian, Yang, Xu, Zhang, Isobe, Hu, Zhang et~al.}]{uni-dpo}
Shangpin Peng, Weinong Wang, Zhuotao Tian, Senqiao Yang, Haotian Xu, Chengquan Zhang, Takashi Isobe, Baotian Hu, Min Zhang, and 1 others.
\newblock Uni-dpo: A unified paradigm for dynamic preference optimization of llms.
\newblock In \emph{The Fourteenth International Conference on Learning Representations}.

\bibitem[{Pi et~al.(2025)Pi, Miao, Li, Liu, Gao, Zhang, and Zhou}]{pi2025pointing}
Renjie Pi, Kehao Miao, Peihang Li, Runtao Liu, Jiahui Gao, Jipeng Zhang, and Xiaofang Zhou. 2025.
\newblock Pointing to a {Llama} and call it a camel: On the sycophancy of multimodal large language models.
\newblock \emph{arXiv preprint arXiv:2509.16149}.

\bibitem[{SeekWorld(2025)}]{seekworld}
SeekWorld. 2025.
\newblock Seekworld: Geolocation is a natural rl task for o3-like visual clue-tracking reasoning.
\newblock \emph{https://github.com/TheEighthDay}.

\bibitem[{Shao et~al.(2024)Shao, Qian, Xiao, Song, Zong, Wang, Liu, and Li}]{visual-cot}
Hao Shao, Shengju Qian, Han Xiao, Guanglu Song, Zhuofan Zong, Letian Wang, Yu~Liu, and Hongsheng Li. 2024.
\newblock Visual cot: Advancing multi-modal language models with a comprehensive dataset and benchmark for chain-of-thought reasoning.
\newblock \emph{Advances in Neural Information Processing Systems}, 37:8612--8642.

\bibitem[{Su et~al.(2026)Su, Gao, Guo, Liu, Zhang, Geng, Huang, Xia, Jiang, Wang et~al.}]{agentvista}
Zhaochen Su, Jincheng Gao, Hangyu Guo, Zhenhua Liu, Lueyang Zhang, Xinyu Geng, Shijue Huang, Peng Xia, Guanyu Jiang, Cheng Wang, and 1 others. 2026.
\newblock Agentvista: Evaluating multimodal agents in ultra-challenging realistic visual scenarios.
\newblock \emph{arXiv preprint arXiv:2602.23166}.

\bibitem[{Wang et~al.(2025{\natexlab{a}})Wang, Li, Huang, Wang, Wang, Zhang, Zheng, Bai, Kang, Feng et~al.}]{treebench}
Haochen Wang, Xiangtai Li, Zilong Huang, Anran Wang, Jiacong Wang, Tao Zhang, Jiani Zheng, Sule Bai, Zijian Kang, Jiashi Feng, and 1 others. 2025{\natexlab{a}}.
\newblock Traceable evidence enhanced visual grounded reasoning: Evaluation and methodology.
\newblock \emph{arXiv preprint arXiv:2507.07999}.

\bibitem[{Wang et~al.(2025{\natexlab{b}})Wang, Su, Ren, Lin, and Chen}]{pixel_reasoner}
Haozhe Wang, Alex Su, Weiming Ren, Fangzhen Lin, and Wenhu Chen. 2025{\natexlab{b}}.
\newblock Pixel reasoner: Incentivizing pixel-space reasoning with curiosity-driven reinforcement learning.
\newblock \emph{arXiv preprint arXiv:2505.15966}.

\bibitem[{Wang et~al.(2024)Wang, Pan, Shi, Lu, Ren, Zhou, Zhan, and Li}]{MathVision}
Ke~Wang, Junting Pan, Weikang Shi, Zimu Lu, Houxing Ren, Aojun Zhou, Mingjie Zhan, and Hongsheng Li. 2024.
\newblock Measuring multimodal mathematical reasoning with math-vision dataset.
\newblock \emph{Advances in Neural Information Processing Systems}, 37:95095--95169.

\bibitem[{Wang et~al.(2025{\natexlab{c}})Wang, Gao, Gu, Pu, Cui, Wei, Liu, Jing, Ye, Shao et~al.}]{internvl3.5}
Weiyun Wang, Zhangwei Gao, Lixin Gu, Hengjun Pu, Long Cui, Xingguang Wei, Zhaoyang Liu, Linglin Jing, Shenglong Ye, Jie Shao, and 1 others. 2025{\natexlab{c}}.
\newblock Internvl3. 5: Advancing open-source multimodal models in versatility, reasoning, and efficiency.
\newblock \emph{arXiv preprint arXiv:2508.18265}.

\bibitem[{Wang et~al.(2025{\natexlab{d}})Wang, Ding, Zeng, Zhou, Shen, Luo, Yu, and Tao}]{hrbench}
Wenbin Wang, Liang Ding, Minyan Zeng, Xiabin Zhou, Li~Shen, Yong Luo, Wei Yu, and Dacheng Tao. 2025{\natexlab{d}}.
\newblock Divide, conquer and combine: A training-free framework for high-resolution image perception in multimodal large language models.
\newblock In \emph{Proceedings of the AAAI Conference on Artificial Intelligence}, volume~39, pages 7907--7915.

\bibitem[{Wu and Xie(2024)}]{v_star}
Penghao Wu and Saining Xie. 2024.
\newblock V$^*$: Guided visual search as a core mechanism in multimodal llms.
\newblock In \emph{Proceedings of the IEEE/CVF Conference on Computer Vision and Pattern Recognition}, pages 13084--13094.

\bibitem[{xAI(2024)}]{realworldqa}
xAI. 2024.
\newblock Realworldqa: A benchmark for real-world spatial understanding.
\newblock \emph{https://huggingface.co/ datasets/xai-org/RealworldQA}.

\bibitem[{Xiang et~al.(2025)Xiang, Li, Zhang, Huang, Liu, Qu, He, Chen, Yuan, Han et~al.}]{seephys}
Kun Xiang, Heng Li, Terry~Jingchen Zhang, Yinya Huang, Zirong Liu, Peixin Qu, Jixi He, Jiaqi Chen, Yu-Jie Yuan, Jianhua Han, and 1 others. 2025.
\newblock Seephys: Does seeing help thinking?--benchmarking vision-based physics reasoning.
\newblock \emph{arXiv preprint arXiv:2505.19099}.

\bibitem[{Xu et~al.(2025)Xu, Sun, Du, Li, Lyu, and Yuan}]{vacot}
Zhengzhuo Xu, Chong Sun, SiNan Du, Chen Li, Jing Lyu, and Chun Yuan. 2025.
\newblock Vacot: Rethinking visual data augmentation with vlms.
\newblock \emph{arXiv preprint arXiv:2512.02361}.

\bibitem[{Yu et~al.(2025)Yu, Chen, Ju, Jia, Zhang, Huang, Wu, Cui, Ran, Zhang et~al.}]{sibench}
Songsong Yu, Yuxin Chen, Hao Ju, Lianjie Jia, Fuxi Zhang, Shaofei Huang, Yuhan Wu, Rundi Cui, Binghao Ran, Zaibin Zhang, and 1 others. 2025.
\newblock How far are vlms from visual spatial intelligence? a benchmark-driven perspective.
\newblock \emph{arXiv preprint arXiv:2509.18905}.

\bibitem[{Yue et~al.(2024)Yue, Ni, Zhang, Zheng, Liu, Zhang, Stevens, Jiang, Ren, Sun et~al.}]{mmmu}
Xiang Yue, Yuansheng Ni, Kai Zhang, Tianyu Zheng, Ruoqi Liu, Ge~Zhang, Samuel Stevens, Dongfu Jiang, Weiming Ren, Yuxuan Sun, and 1 others. 2024.
\newblock Mmmu: A massive multi-discipline multimodal understanding and reasoning benchmark for expert agi.
\newblock In \emph{Proceedings of the IEEE/CVF Conference on Computer Vision and Pattern Recognition}, pages 9556--9567.

\bibitem[{Yue et~al.(2025)Yue, Zheng, Ni, Wang, Zhang, Tong, Sun, Yu, Zhang, Sun et~al.}]{mmmu-pro}
Xiang Yue, Tianyu Zheng, Yuansheng Ni, Yubo Wang, Kai Zhang, Shengbang Tong, Yuxuan Sun, Botao Yu, Ge~Zhang, Huan Sun, and 1 others. 2025.
\newblock Mmmu-pro: A more robust multi-discipline multimodal understanding benchmark.
\newblock In \emph{Proceedings of the 63rd Annual Meeting of the Association for Computational Linguistics (Volume 1: Long Papers)}, pages 15134--15186.

\bibitem[{Zhang et~al.(2025{\natexlab{a}})Zhang, Yao, Pi, Liang, and Fung}]{vlm2-bench}
Jianshu Zhang, Dongyu Yao, Renjie Pi, Paul~Pu Liang, and Yi~R Fung. 2025{\natexlab{a}}.
\newblock Vlm2-bench: A closer look at how well vlms implicitly link explicit matching visual cues.
\newblock \emph{arXiv preprint arXiv:2502.12084}.

\bibitem[{Zhang et~al.(2024{\natexlab{a}})Zhang, Jiang, Zhang, Lin, Guo, Qiu, Zhou, Lu, Chang, Qiao et~al.}]{mathverse}
Renrui Zhang, Dongzhi Jiang, Yichi Zhang, Haokun Lin, Ziyu Guo, Pengshuo Qiu, Aojun Zhou, Pan Lu, Kai-Wei Chang, Yu~Qiao, and 1 others. 2024{\natexlab{a}}.
\newblock Mathverse: Does your multi-modal llm truly see the diagrams in visual math problems?
\newblock In \emph{European Conference on Computer Vision}, pages 169--186. Springer.

\bibitem[{Zhang et~al.(2025{\natexlab{b}})Zhang, Lu, Yin, Fu, Chen, Hu, Wen, Jiang, Liu, Zhang et~al.}]{thyme}
Yi-Fan Zhang, Xingyu Lu, Shukang Yin, Chaoyou Fu, Wei Chen, Xiao Hu, Bin Wen, Kaiyu Jiang, Changyi Liu, Tianke Zhang, and 1 others. 2025{\natexlab{b}}.
\newblock Thyme: Think beyond images.
\newblock \emph{arXiv preprint arXiv:2508.11630}.

\bibitem[{Zhang et~al.(2024{\natexlab{b}})Zhang, Zhang, Tian, Fu, Zhang, Wu, Li, Wang, Wen, Zhang et~al.}]{mme-realworld}
Yi-Fan Zhang, Huanyu Zhang, Haochen Tian, Chaoyou Fu, Shuangqing Zhang, Junfei Wu, Feng Li, Kun Wang, Qingsong Wen, Zhang Zhang, and 1 others. 2024{\natexlab{b}}.
\newblock Mme-realworld: Could your multimodal llm challenge high-resolution real-world scenarios that are difficult for humans?
\newblock \emph{arXiv preprint arXiv:2408.13257}.

\bibitem[{Zhao et~al.(2025)Zhao, Zhang, Lin, Li, Wu, Zhang, and Wei}]{pyvision}
Shitian Zhao, Haoquan Zhang, Shaoheng Lin, Ming Li, Qilong Wu, Kaipeng Zhang, and Chen Wei. 2025.
\newblock Pyvision: Agentic vision with dynamic tooling.
\newblock \emph{arXiv preprint arXiv:2507.07998}.

\bibitem[{Zheng et~al.(2025{\natexlab{a}})Zheng, Wu, Wang, and Jiang}]{zheng2025unveiling}
Xinhan Zheng, Huyu Wu, Xueting Wang, and Haiyun Jiang. 2025{\natexlab{a}}.
\newblock Unveiling intrinsic text bias in multimodal large language models through attention key-space analysis.
\newblock \emph{arXiv preprint arXiv:2510.26721}.

\bibitem[{Zheng et~al.(2025{\natexlab{b}})Zheng, Yang, Hong, Zhao, Xu, Yang, Shen, and Yu}]{deepeyes}
Ziwei Zheng, Michael Yang, Jack Hong, Chenxiao Zhao, Guohai Xu, Le~Yang, Chao Shen, and Xing Yu. 2025{\natexlab{b}}.
\newblock Deepeyes: Incentivizing" thinking with images" via reinforcement learning.
\newblock \emph{arXiv preprint arXiv:2505.14362}.

\bibitem[{Zhou et~al.(2025)Zhou, Chen, Ma, Hu, Fu, Wang, Wan, Zhao, and Krishna}]{revpt}
Zetong Zhou, Dongping Chen, Zixian Ma, Zhihan Hu, Mingyang Fu, Sinan Wang, Yao Wan, Zhou Zhao, and Ranjay Krishna. 2025.
\newblock Reinforced visual perception with tools.
\newblock \emph{arXiv preprint arXiv:2509.01656}.

\bibitem[{Zhu et~al.(2025)Zhu, Wang, Chen, Liu, Ye, Gu, Tian, Duan, Su, Shao et~al.}]{internvl3}
Jinguo Zhu, Weiyun Wang, Zhe Chen, Zhaoyang Liu, Shenglong Ye, Lixin Gu, Hao Tian, Yuchen Duan, Weijie Su, Jie Shao, and 1 others. 2025.
\newblock Internvl3: Exploring advanced training and test-time recipes for open-source multimodal models.
\newblock \emph{arXiv preprint arXiv:2504.10479}.

\end{thebibliography}

\appendix
\renewcommand\thesection{\Alph{section}}
\setcounter{equation}{0}
\setcounter{section}{0}
\section{Use of AI Assistant}
\label{sec:appendix_aiuseage}
We incorporate GPT-5~\cite{gpt5} to assist with code writing, specifically for data processing and evaluation scripts. Additionally, the model is employed to proofread and correct grammatical errors throughout this paper.

\section{Accuracy Calculation}
\label{sec:appendix_}
For each task, accuracy is calculated as the ratio of correctly answered questions to the total number of questions within this task. The overall accuracy is computed as the total number of correct predictions across the entire benchmark divided by the total count of questions. The results reported in Table~\ref{tab:main_result} represent the average of three independent runs.

Formally, let $N_{correct}^{(i)}$ and $N_{total}^{(i)}$ denote the number of correctly answered questions and the total number of questions for task $i$, respectively. The accuracy for task $i$, denoted as $\text{Acc}_i$, and the overall accuracy, $\text{Acc}_{\text{overall}}$, are calculated as follows:

\begin{equation}
    \text{Acc}_i = \frac{N_{correct}^{(i)}}{N_{total}^{(i)}}
\end{equation}

\begin{equation}
    \text{Acc}_{\text{overall}} = \frac{\sum_{i} N_{correct}^{(i)}}{\sum_{i} N_{total}^{(i)}}
\end{equation}

\section{System Prompt}
\label{sec:appendix_system_prompt}
\begin{figure*}[t]
    \centering
    \includegraphics[width=0.95\linewidth]{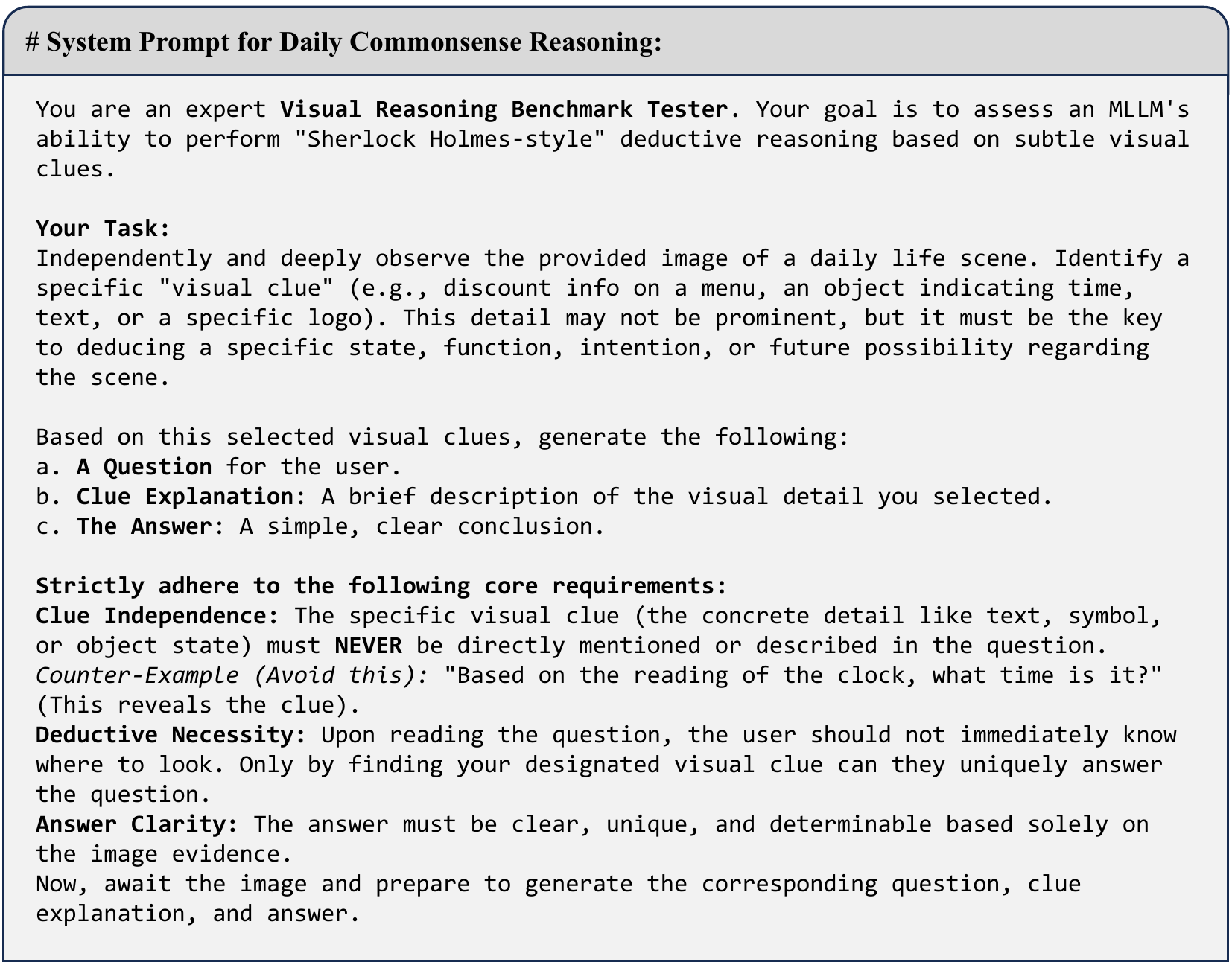}
    \caption{System prompt for constructing question-clue-answer triplets in Daily Commonsense Reasoning.}
    \label{fig:appendix_system_prompt_daily}
\end{figure*}

\begin{figure*}[t]
    \centering
    \includegraphics[width=0.95\linewidth]{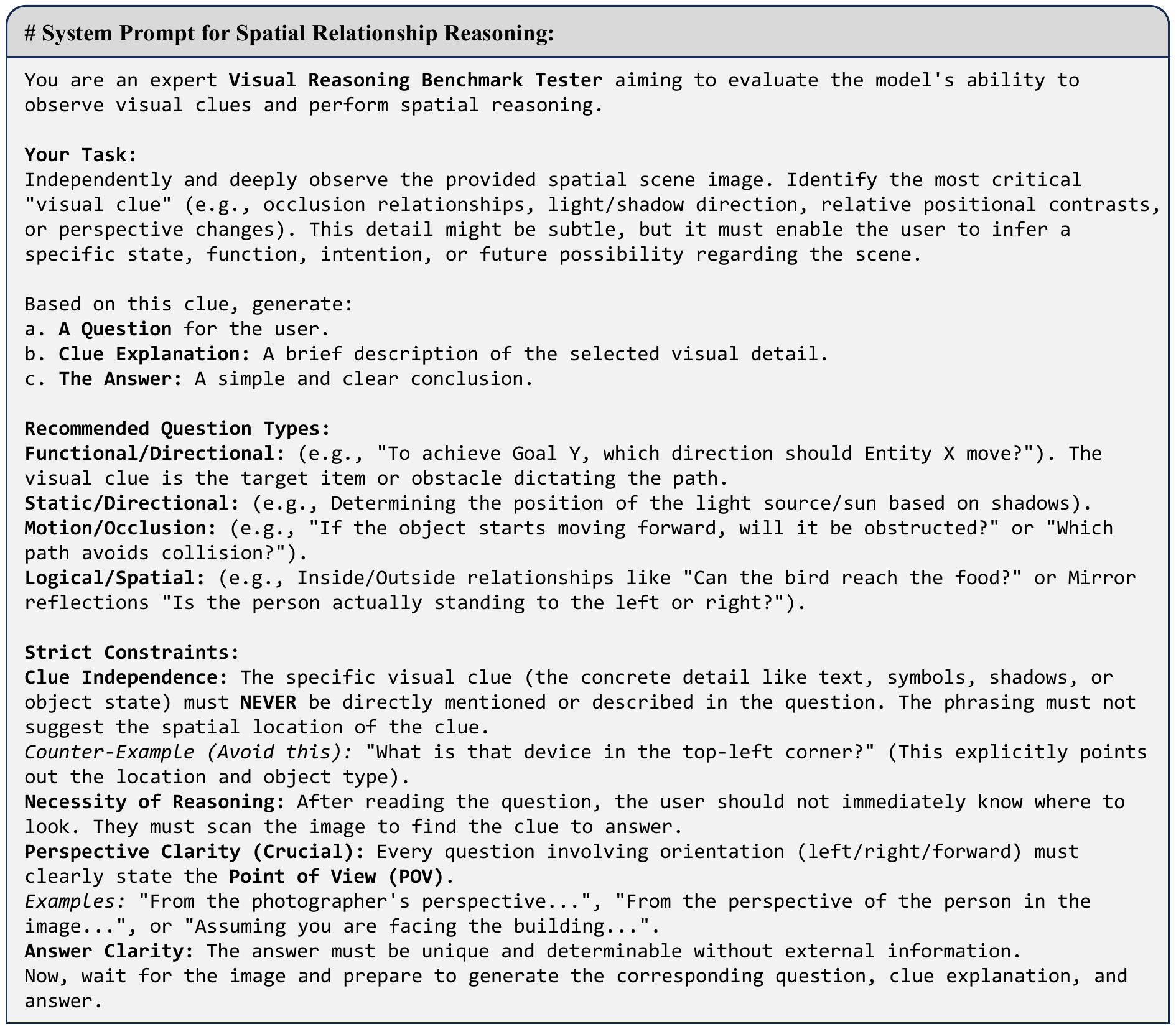}
    \caption{System prompt for constructing question-clue-answer triplets in Spatial Relationship Reasoning.}
    \label{fig:appendix_system_prompt_spatial}
\end{figure*}

\begin{figure*}[t]
    \centering
    \includegraphics[width=0.95\linewidth]{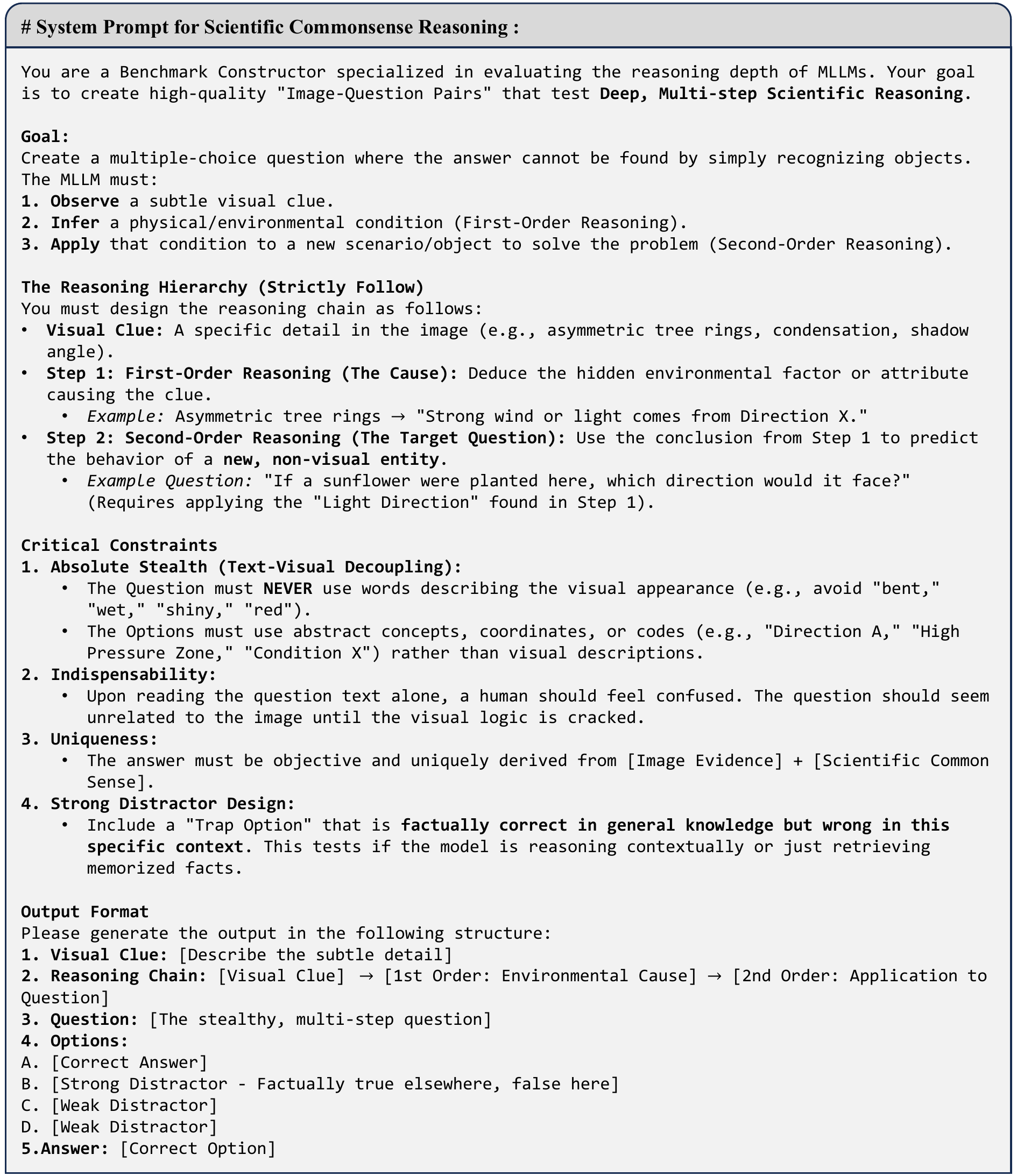}
    \caption{System prompt for constructing question-clue-answer triplets in Scientific Commonsense Reasoning.}
    \label{fig:appendix_system_prompt_science}
\end{figure*}

\paragraph{System prompt for constructing triplet.}
The system prompts designed to generate triplets for Daily Commonsense, Spatial Relationship Reasoning, and Scientific Commonsense are detailed in Figures~\ref{fig:appendix_system_prompt_daily}--\ref{fig:appendix_system_prompt_science}.
For Location Identification, we use a fixed prompt: "In which country and within which first-level administrative region of that country was this picture taken?". Given that the ground-truth locations are verified during data collection, the model's role (Gemini-2.5-Pro) is limited to extracting and generating visual clues.

\begin{figure}[t]
    \centering
    \includegraphics[width=0.48\textwidth]{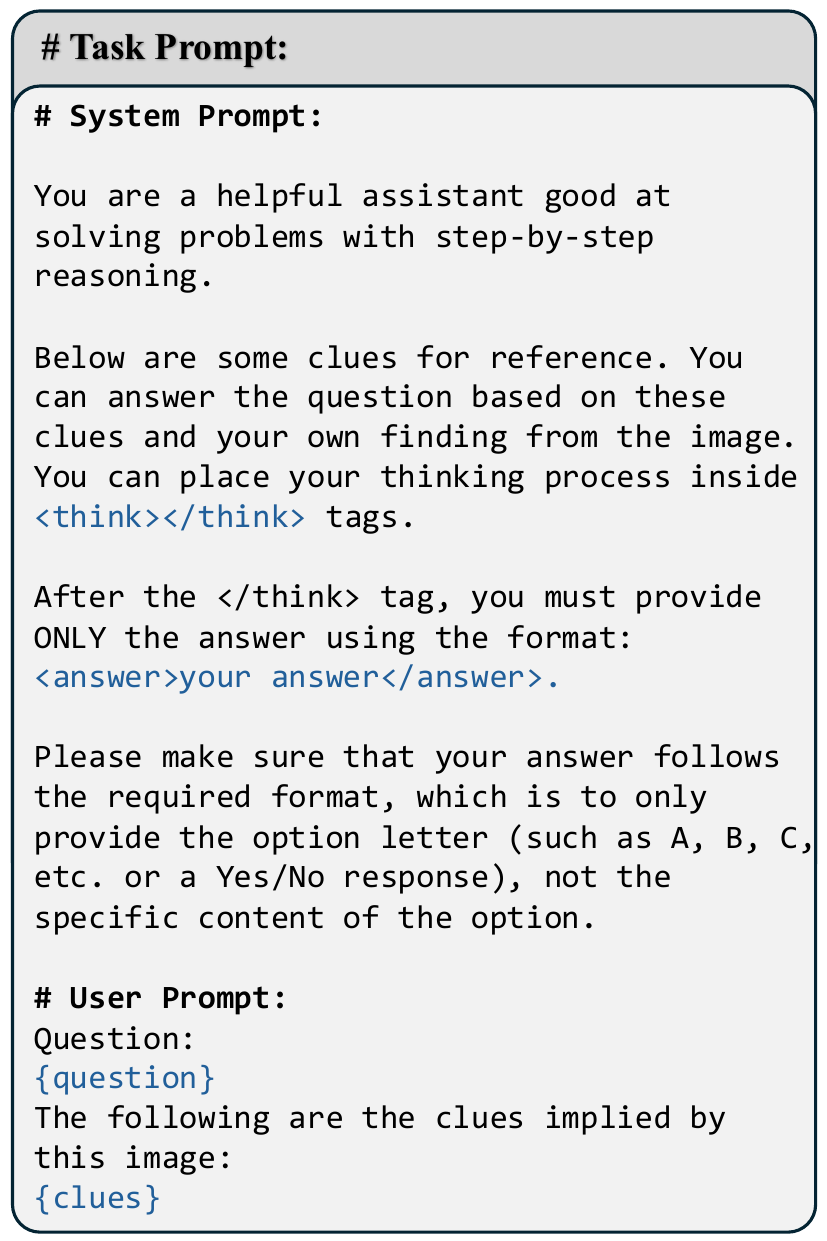}
    \caption{System and user prompts used for injecting visual clues from external models during inference.}
    \label{fig:appendix_sys_prompt_exchange_clue}
\end{figure}

\begin{figure}[t]
    \centering
    \includegraphics[width=0.48\textwidth]{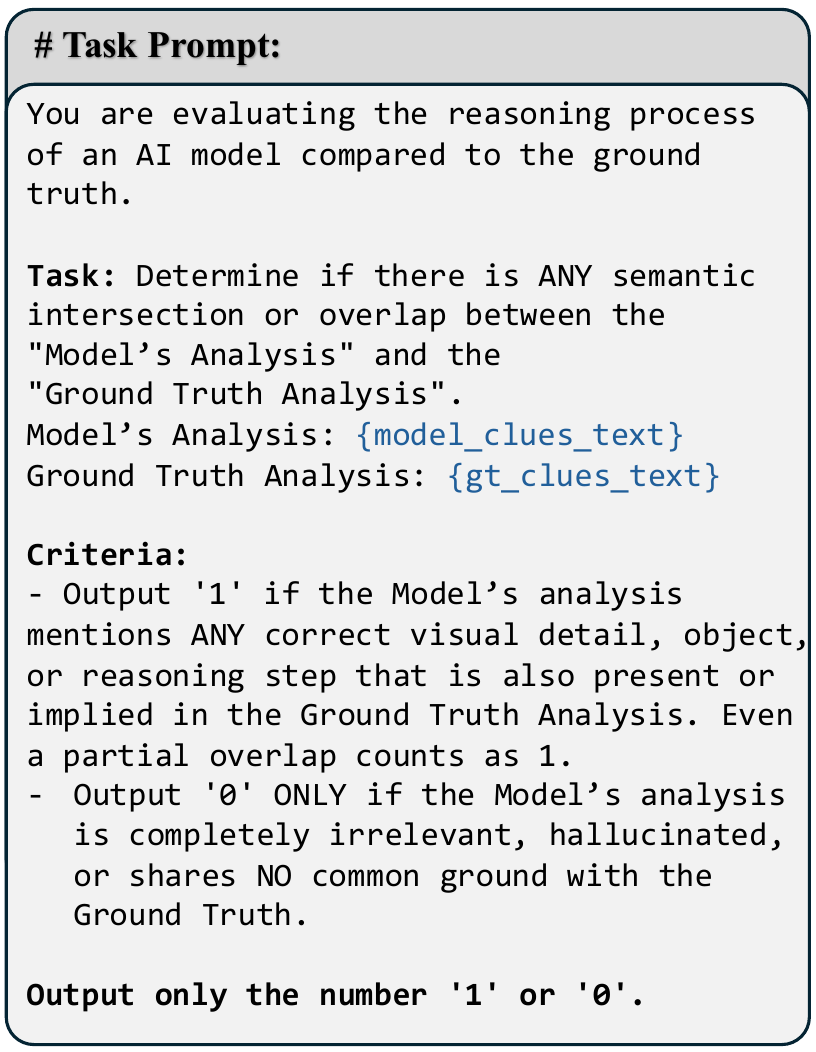}
    \caption{System prompt used for rigorous evaluation, with Gemini-2.5-Pro serving as the judge model. }
    \label{fig:appendix_rigorous_eval_prompt}
\end{figure}

\paragraph{System and user prompts for injecting external clues.}
Figure~\ref{fig:appendix_sys_prompt_exchange_clue} illustrates the system and user prompts employed to enable the model to reference visual clues generated by external models. Specifically, the system prompt instructs the model to utilize these external clues, while the specific clues are injected via the user prompt.

\paragraph{System prompt used for rigorous evaluation.}
Figure~\ref{fig:appendix_rigorous_eval_prompt} shows the prompt for the LLM-based judge.
Its primary role is to assess whether the predicted clues semantically match the ground truth. This verification confirms that the model's correct predictions are supported by reasonable evidence, effectively distinguishing actual reasoning from lucky guesses.

\begin{table}[t]
\renewcommand\arraystretch{1.1}
\centering
\caption{
Participant assignment for the Human Baseline evaluation. Participants A--F represent six distinct undergraduate students. The assignment ensures that each questionnaire category is completed three times.
}
\label{tab:human_baseline}
\resizebox{0.48\textwidth}{!}{%
\fontsize{26pt}{30pt}\selectfont
\begin{tabular}{lccccc}
\toprule
\textbf{Participant} & \textbf{\makecell{Location \\ Identification}} & \textbf{\makecell{Spatial \\ Relationship}} & \textbf{\makecell{Daily  \\ Commonsense}} & \textbf{\makecell{Scientific \\ Commonsense}}  & \textbf{Total} \\ 
\midrule
A & \textcolor{nvidiagreen}{\ding{51}} & \textcolor{nvidiagreen}{\ding{51}} & - & - & 2 \\
B & \textcolor{nvidiagreen}{\ding{51}} & - & \textcolor{nvidiagreen}{\ding{51}} & - & 2 \\
C & \textcolor{nvidiagreen}{\ding{51}} & - & - & \textcolor{nvidiagreen}{\ding{51}} & 2 \\
D & - & \textcolor{nvidiagreen}{\ding{51}} & \textcolor{nvidiagreen}{\ding{51}} & - & 2 \\
E & - & \textcolor{nvidiagreen}{\ding{51}} & - & \textcolor{nvidiagreen}{\ding{51}} & 2 \\ 
F & - & - & \textcolor{nvidiagreen}{\ding{51}} & \textcolor{nvidiagreen}{\ding{51}} & 2 \\
\midrule
Total & 3 & 3 & 3 & 3 & 12 \\
\bottomrule
\end{tabular}
}
\end{table}

\section{More Experiment Details}
\subsection{Experiment Setup}
To ensure a fair comparison, we standardize experimental configurations across all models. Open-source models with fewer than 10B parameters are evaluated using 1--2 NVIDIA A800~(80GB) GPUs. Conversely, larger models utilize 4--8 NVIDIA H20~(96GB) GPUs to accommodate their higher memory and computational demands.

\subsection{Human Baseline}
\label{sec:human_baseline}
We recruit six undergraduate participants to complete a total of 12 questionnaires. The study follows a balanced design where each participant completes two questionnaires, ensuring that every category is covered by three independent responses.

To ensure the validity of the human baseline, participants are provided with strict guidelines:
\begin{itemize}
    \item Diligent answering: Participants are required to answer 50 questions per questionnaire carefully, performing necessary calculations rather than guessing randomly.
    \item Active visual exploration: They are explicitly instructed to actively identify visual clues within the images to assist in reasoning.
    \item Tool restrictions: The use of LLMs or reverse image search tools (e.g., Google Chrome) is strictly prohibited to accurately benchmark the gap between human and MLLM performance.
    \item Permissible search: Limited text-based web searches are allowed solely for recalling specific facts (e.g., verifying the location of a recognized landmark or clarifying scientific terminology) but not for direct problem-solving.
\end{itemize}

\subsection{Textual Bias and Visual Sycophancy}
\label{sec:textual_bias}
\citet{zheng2025unveiling} reveals that MLLMs exhibit a structural bias toward textual inputs. Furthermore, \citet{pi2025pointing} identifies a prevalent ``visual sycophancy'' behavior, where the model's visual judgment is heavily influenced by concurrent textual conditions. This dependency means the model may override its own visual perception to match the clues, making the final inference highly sensitive to the quality of the injected text.

\section{More Experimental Results}
\label{sec:more_expr_results}
\begin{figure*}[t]
    \centering
    \includegraphics[width=0.90\textwidth]{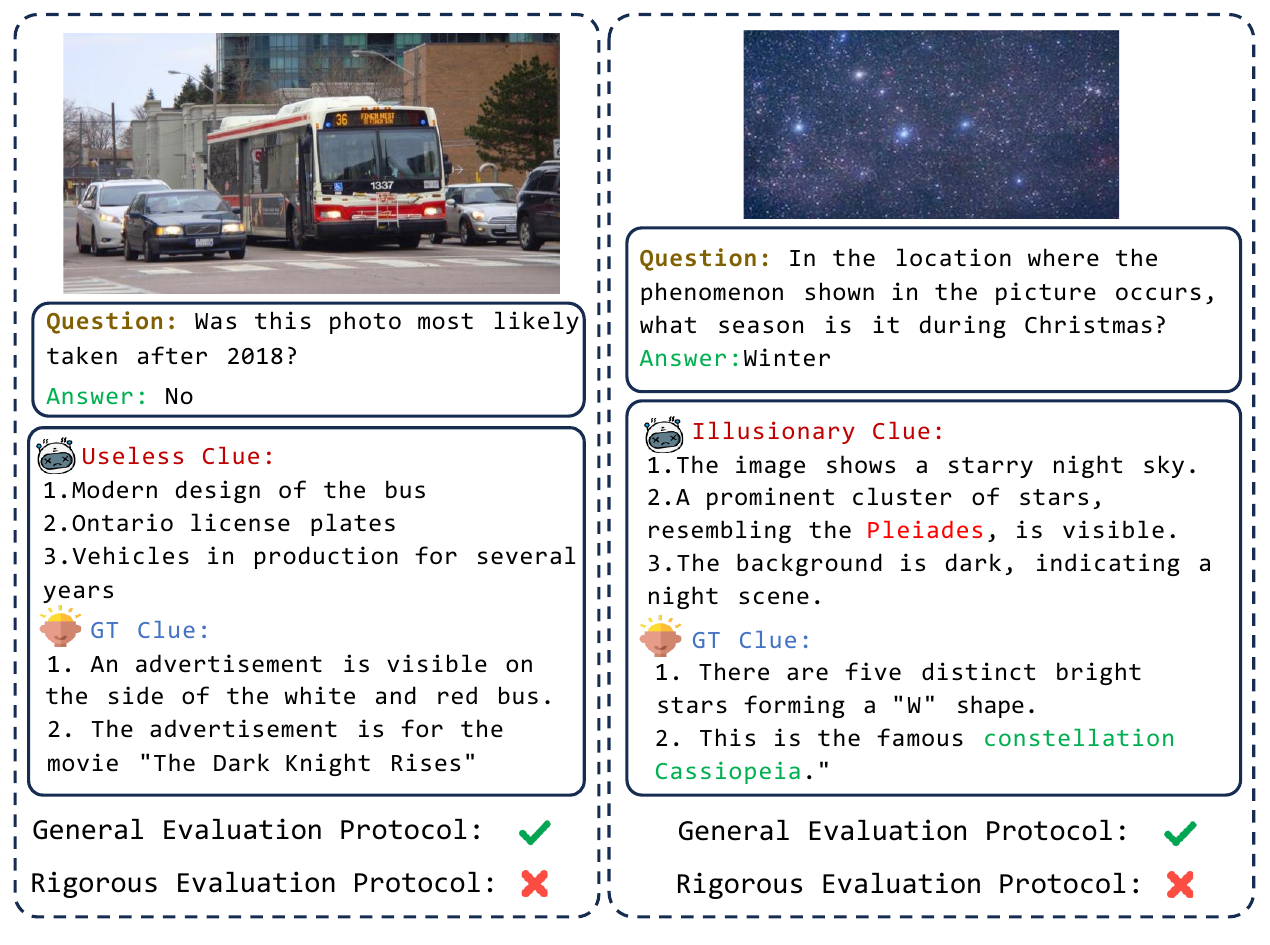}
    \caption{Qualitative visualization of visual clues under the Rigorous Evaluation Protocol.}
    \label{fig:appendix_rigorous_viz}
\end{figure*}

\paragraph{Qualitatively results under Rigorous Evaluation.}
In Figure~\ref{fig:appendix_rigorous_viz}, we qualitatively present instances where model responses are deemed correct under the General Evaluation Protocol but fail under the Rigorous Evaluation Protocol. This discrepancy primarily stems from the model's reliance on useless or illusionary clues during reasoning, exposing behaviors of \textit{lucky guesses}. This further underscores that Rigorous Evaluation Protocol is more rigorous than the General Evaluation Protocol.

\begin{figure*}[t]
    \centering
    \includegraphics[width=0.95\linewidth]{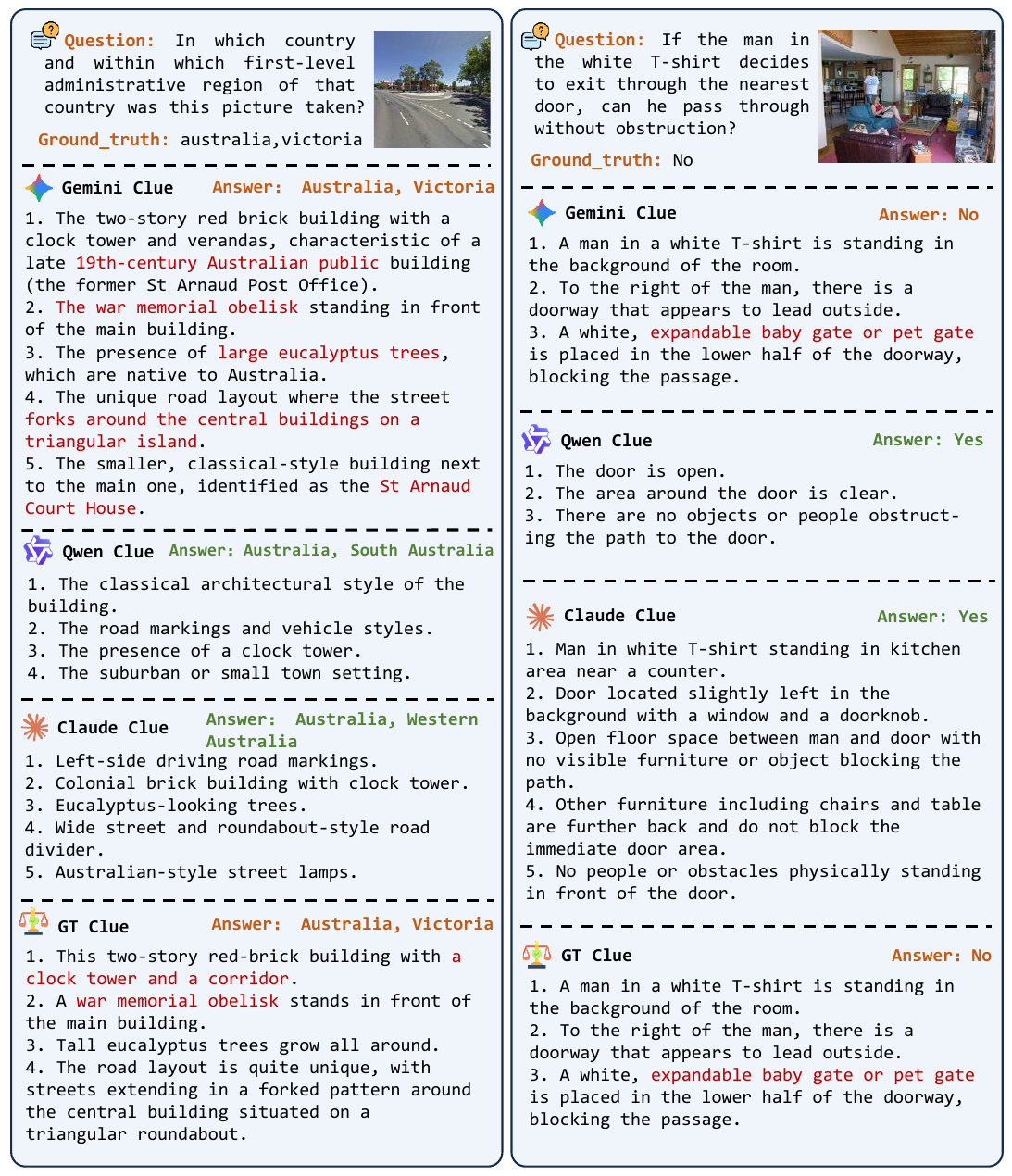}
    \caption{Comparison of answer generation under different clue contexts. We feed Claude-3.7 with visual clues from varying sources as additional context. }
    \label{fig:appendix_exchange_clue}
\end{figure*}

\paragraph{Qualitative analysis of external clue injection.}
We also present additional comparisons involving the injection of external visual clues in Figure~\ref{fig:appendix_exchange_clue}. The visualizations demonstrate that utilizing correct visual clues effectively improves the model's reasoning accuracy.

\begin{figure*}[t]
    \centering
    \includegraphics[width=0.95\linewidth]{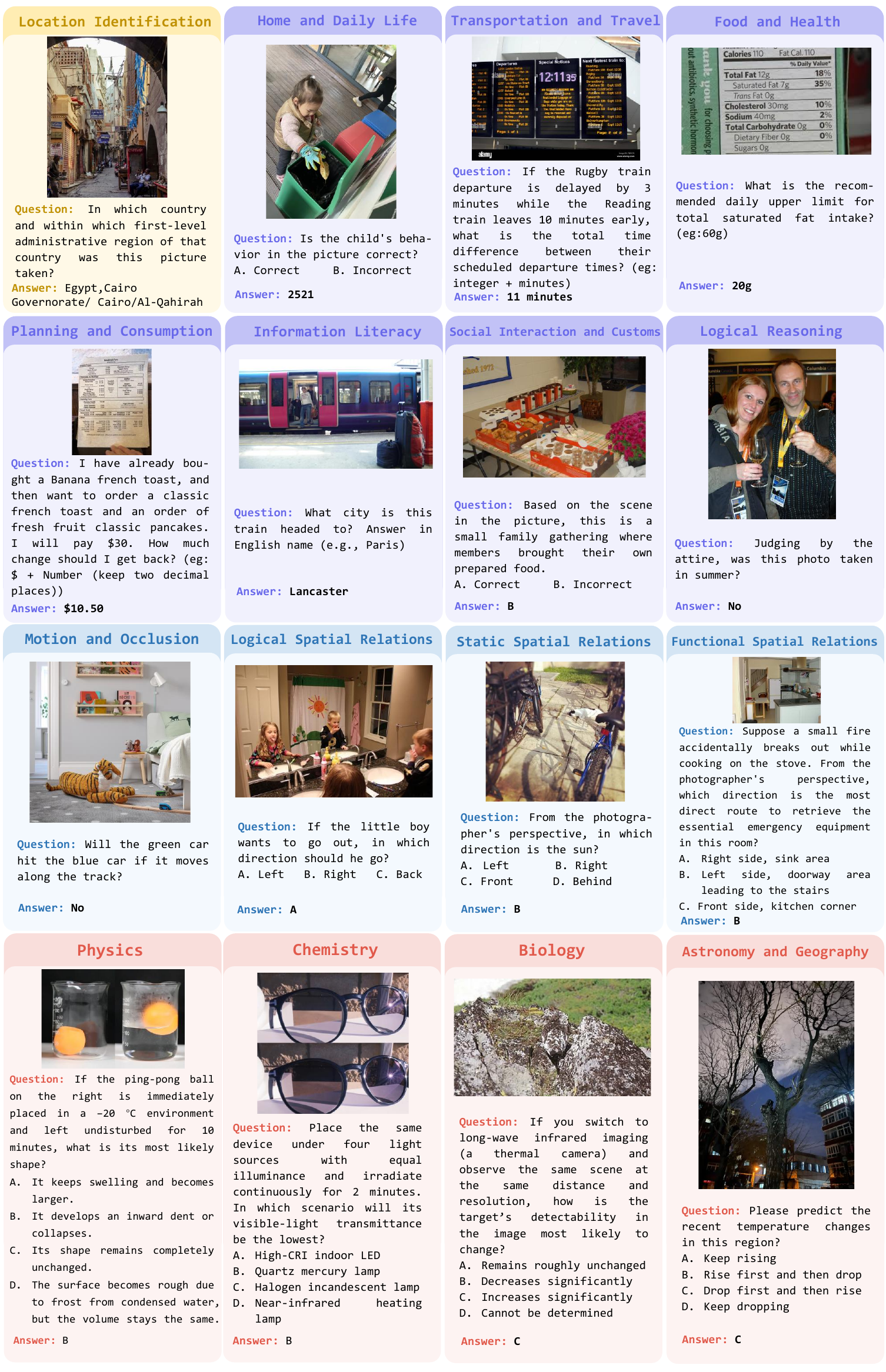}
    \caption{Illustrative examples of the 16 subtasks, with four colors representing four scenarios.}
    \label{fig:appendix_example_viz}
\end{figure*}

\end{document}